\documentclass[10pt,twocolumn,letterpaper]{article}
\pdfoutput=1
\usepackage[pagenumbers]{cvpr} 
\usepackage{algorithm}
\usepackage{algorithmic}

\usepackage{booktabs}
\usepackage{pifont}
\usepackage[table,x11names, dvipsnames]{xcolor}

\newlength\savewidth
\newcommand\paperurl[1]{{\footnotesize{\color{blue}{\url{#1}}}}}

\usepackage{url}            
\usepackage{booktabs}       
\usepackage{amsfonts}       
\usepackage{nicefrac}       
\usepackage{microtype}      

\usepackage[dvipsnames]{xcolor}
\newcommand{\shortname}{LLaVA-G}
\newcommand{\fullname}{LLaVA-Grounding}
\newcommand{\benchname}{Grounding-Bench}

\definecolor{green1}{RGB}{104,164,68}
\definecolor{orange1}{RGB}{241, 194, 67}

\newcommand{\VarSty}[1]{\textnormal{\ttfamily\color{blue!90!black}#1}\unskip}
\definecolor{mygreen}{HTML}{3cb44b}
\newcommand{\PredSty}[1]
{\textnormal{\ttfamily\color{mygreen!90!black}#1}\unskip}
\usepackage{tcolorbox}
\usepackage{multirow}
\usepackage{color, colortbl}
\usepackage{duckuments}
\usepackage[pagebackref,breaklinks,colorlinks,citecolor=cvprblue]{hyperref}
\usepackage{graphicx}
\usepackage{xcolor}
\usepackage[normalem]{ulem}
\useunder{\uline}{\ul}{}
\usepackage{multirow}
\usepackage{wrapfig}
\usepackage{pifont}
\usepackage[T1]{fontenc}
\usepackage{algorithm}
\usepackage{algorithmic}
\usepackage[algo2e]{algorithm2e}

\SetKwComment{Comment}{\color{green!50!black}\# }{}

\SetKwProg{Function}{def}{:}{}

\SetKwProg{For}{for}{:}{}
\SetKwProg{If}{if}{:}{}
\definecolor{people}{rgb}{0.666667,0,1}
\definecolor{laptop}{rgb}{1,0.666667,0}
\definecolor{TV}{rgb}{0,1,0}
\definecolor{clock}{rgb}{0,1,0.662745}
\definecolor{books}{rgb}{0.666667,1,0}
\definecolor{cell_phone}{rgb}{1,0,0}
\definecolor{chair}{rgb}{1,0,0.666667}
\definecolor{bottle}{rgb}{0,0,1}
\definecolor{color_182125}{rgb}{0.6,0.670588,0}
\definecolor{color_35381}{rgb}{0,0.686275,0.647059}
\definecolor{color_193416}{rgb}{0.666667,0,1}
\definecolor{color_261472}{rgb}{0.929412,0.490196,0.192157}
\definecolor{color_122260}{rgb}{0.356863,0.607843,0.835294}
\definecolor{color_142857}{rgb}{0.439216,0.678431,0.278431}
\definecolor{color_139071}{rgb}{0.439216,0.188235,0.627451}
\definecolor{color_214303}{rgb}{0.752941,0,0}
\definecolor{color_282751}{rgb}{1,1,0}
\definecolor{color_274846}{rgb}{1,0,0}
\definecolor{color_37696}{rgb}{0,1,0}
\definecolor{color_30046}{rgb}{0,0,1}
\definecolor{color_238111}{rgb}{0.823529,0.823529,0}
\definecolor{color_29791}{rgb}{0,0,0}
\definecolor{color_37951}{rgb}{0,1,1}
\definecolor{color_159743}{rgb}{0.498039,1,0}
\definecolor{color_152093}{rgb}{0.498039,0,1}
\definecolor{color_275101}{rgb}{1,0,1}
\definecolor{color_182125}{rgb}{0.6,0.670588,0}
\definecolor{color_29791}{rgb}{0,0,0}
\definecolor{color_35381}{rgb}{0,0.686275,0.647059}
\definecolor{color_88082}{rgb}{0.215686,0.662745,0.772549}
\definecolor{color_229830}{rgb}{0.8,0.47451,0.956863}
\definecolor{color_60908}{rgb}{0.121569,0.160784,0.215686}
\definecolor{color_91812}{rgb}{0.231373,0.643137,0.933333}

\definecolor{cvprblue}{rgb}{0.21,0.49,0.74}
\usepackage[pagebackref,breaklinks,colorlinks,citecolor=cvprblue]{hyperref}

\usepackage{amsmath}
\usepackage{amsfonts}
\usepackage{amssymb}
\usepackage{wrapfig}
\usepackage{subcaption}
\usepackage{multirow}
 \usepackage{mathtools} 

\usepackage{verbatim}

\usepackage{anyfontsize}

\usepackage{microtype}
\usepackage{graphicx}
\usepackage{booktabs} 
\usepackage{multirow}
\usepackage{amsmath,amssymb}
\usepackage{booktabs}
\usepackage{caption,subcaption}

\usepackage{xcolor}
\definecolor{mygreen}{HTML}{3cb44b}
\definecolor{skyblue}{HTML}{beffff}
\definecolor{lightgreen}{HTML}{90ee90}

\usepackage{color, colortbl}

\definecolor{emerald}{rgb}{0.31, 0.78, 0.37}

\usepackage{tcolorbox}
\usepackage{enumitem}
\setitemize{itemsep=10pt,topsep=0pt,parsep=0pt,partopsep=0pt}

\usepackage{colortbl}

\usepackage{xcolor}
\definecolor{mygreen}{HTML}{3cb44b}
\colorlet{myyellow}{green!10!orange!90!}
\makeatletter

\usepackage{tikz}
\usetikzlibrary{arrows,shapes,snakes,automata,backgrounds,fit,petri}
\usepackage{adjustbox}

\newcommand{\RN}[1]{%
	\textup{\lowercase\expandafter{\it \romannumeral#1}}%
}
\usepackage{tabu}

\newcommand{\beq}{\vspace{0mm}\begin{equation}}
\newcommand{\eeq}{\vspace{0mm}\end{equation}}
\newcommand{\beqs}{\vspace{0mm}\begin{eqnarray}}
\newcommand{\eeqs}{\vspace{0mm}\end{eqnarray}}
\newcommand{\barr}{\begin{array}}
\newcommand{\earr}{\end{array}}

\newcommand{\Hmat}{{\bf H}}

\newcommand{\Wmat}[0]{{{\bf W}}}
\newcommand{\Xmat}[0]{{{\bf X}}}
\newcommand{\Zmat}[0]{{{\bf X}}}

\usepackage{color, colortbl}
\definecolor{Gray}{gray}{0.93}

\usepackage{lipsum}

\usepackage{pifont}

\usepackage{makecell}

\usepackage{xcolor,amsmath}
\DontPrintSemicolon

\usepackage{xcolor}
\definecolor{mygreen}{HTML}{3cb44b}

\SetKwComment{Comment}{\color{green!50!black}\# }{}

\SetKwProg{Function}{def}{:}{}

\SetKwProg{For}{for}{:}{}
\SetKwProg{If}{if}{:}{}

\def\paperID{5335} 
\def\confName{CVPR}
\def\confYear{2024}

\title{\fullname{}: Grounded Visual Chat with Large Multimodal Models}

\author{
  ~~Hao Zhang$^{\spadesuit*3}$,~~Hongyang Li$^{\diamondsuit*}$, ~~Feng Li$^{\spadesuit3}$, ~~Tianhe Ren$^{\dagger}$, ~~Xueyan Zou$^{\S 3}$,~~Shilong Liu$^{\P 3}$, \\~~Shijia Huang$^{\sharp}$,~~Jianfeng Gao$^{\ddagger2}$,~~Lei Zhang$^{\dagger2}$,~~Chunyuan Li$^{\ddagger1}$,~~Jianwei Yang$^{\ddagger1}$
\and
{
\footnotesize
$^{\spadesuit}$ HKUST \;
$^{\diamondsuit}$ SCUT \;
$^\ddagger$ Microsoft Research, Redmond \;  
$^\dagger$ IDEA \;  
$^{\S}$ UW-Madison \;
$^{\P}$ Tsinghua \;
$^{\sharp}$ CUHK \;
}
\and
\scriptsize{
$^*$~Equal Contribution \;
$1.$~Directional Lead \;
$2.$~Equal Advisory Contribution \;
$3.$~Work performed during an internship at Microsoft \;
}
\\
 \textcolor{magenta}{\texttt{\url{https://llava-vl.github.io/llava-grounding/}}}
}

\begin{document}
\maketitle
\begin{abstract}
With the recent significant advancements in large multimodal models (LMMs), the importance of their grounding capability in visual chat is increasingly recognized. Despite recent efforts to enable LMMs to support grounding, their capabilities for grounding and chat are usually separate, and their chat performance drops dramatically when asked to ground. The problem is the lack of a dataset for grounded visual chat (GVC). Existing grounding datasets only contain short captions. To address this issue, we have created GVC data that allows for the combination of grounding and chat capabilities. To better evaluate the GVC capabilities, we have introduced a benchmark called \benchname{}. Additionally, we have proposed a model design that can support GVC and various types of visual prompts by connecting segmentation models with language models. Experimental results demonstrate that our model outperforms other LMMs on \benchname{}. Furthermore, our model achieves competitive performance on classic grounding benchmarks like RefCOCO/+/g and Flickr30K Entities.
\end{abstract}

\section{Introduction}
\label{sec:intro}
With the success of large language models (LLMs) like GPT-4~\cite{gpt4} and the open-sourced substitutes LLaMA~\cite{touvron2023llama}, researchers are eager to leverage their strong language capabilities in the field of vision. This enthusiasm has led to a surge in the development of large multimodal models (LLMs). Previous LMMs, such as LLaVA~\cite{liu2023visual} and miniGPT-4~\cite{zhu2023minigpt}, have demonstrated exceptional visual chat abilities by generating plausible responses based on images and user instructions. However, they often encounter challenges in providing responses that exhibit a fine-grained understanding of images, including specific regions and alignment with related image regions—this is often referred to as visual grounding.


\begin{figure}[t]
\begin{minipage}{1.0\linewidth}
  \centering
  \includegraphics[width=1.0\linewidth]{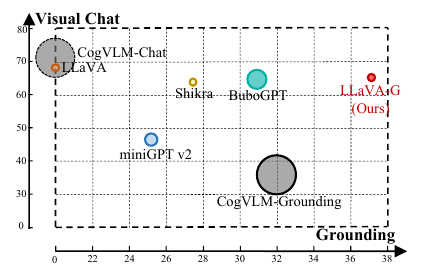}
  \vspace{-4mm}
  \caption{\footnotesize A comparison on the integrated ability of visual grounding and visual chat of open-source LMMs on \benchname. \shortname{} achieves a good trade-off on both abilities simultaneously.  For CogVLM~\cite{wang2023cogvlm}, two different model checkpoints are released: CogVLM-Grounding is the grounding model and CogVLM-Chat is the chat model. Grounding and Visual Chat scores represent the $F_1$ score and Chat scores of detailed descriptions in Table~\ref{tab: our bench new}, respectively. Circle size indicates the model size. }
  \label{fig: teaser}
\end{minipage}
\vspace{-4mm}
\end{figure}

Recognizing the significance of visual grounding for LMMs, recent research efforts have focused on developing grounding and referring capabilities for LMMs~\cite{chen2023shikra,chen2023minigptv2,wang2023cogvlm,you2023ferret,lai2023lisa}. While these models have achieved performance comparable to specialized models~\cite{liu2023grounding,liu2023polyformer} on classic grounding benchmarks such as RefCOCO~\cite{kazemzadeh2014referitgame} and Flickr30K~\cite{plummer2015flickr30k}, they often treat grounding as a distinct task that requires customized prompts to initiate. Consequently, their text responses undergo significant changes when tasked with grounding. Most models, such as MiniGPT-v2~\cite{chen2023minigptv2} and CogVLM-Grounding~\cite{wang2023cogvlm}, can only generate short captions when performing grounding, as they are primarily trained on grounding caption data like Flickr30K. As illustrated in Fig.\ref{fig: teaser}, these earlier models struggle to excel simultaneously in both chat and grounding tasks. BuboGPT\cite{zhao2023bubogpt} maintains chat capability by leveraging an external grounding model for grounding, but this approach can be constrained by the performance of the language encoder in the grounding model. Shikra~\cite{chen2023shikra} engages in referential dialog, which includes grounded chat, but its performance is limited due to the scarcity of available data. All existing LMMs~\cite{chen2023shikra,chen2023minigptv2,you2023ferret,wang2023cogvlm} only support outputting coordinates as text, which restricts localization performance, and they do not support pixel-wise grounding and referring. In summary, previous LMMs struggle to perform grounded visual chat effectively due to the scarcity of grounded visual chat data and suboptimal model designs. Furthermore, they lack the capability for pixel-wise grounding and referring.

To address these challenges, we contribute to grounded visual chat in three key areas: data creation, network architecture, and benchmarking. When annotating grounding data, previous methods such as Kosmos-2~\cite{peng2023kosmos} and GPT4ROI~\cite{zhang2023gpt4roi} rely on pretrained grounding models or detection models to predict bounding boxes based on existing captions. In contrast, we label grounded visual chat data using human-labeled object detection data~\cite{lin2014microsoft}.

Our data creation process begins by leveraging GPT-4~\cite{gpt4}, following the data creation method used in LLaVA~\cite{liu2023visual}. We provide GPT-4 with chat data and ground-truth instances, instructing it to match instances with noun phrases in the chat data. This approach benefits from the high quality of human-labeled instances and chat data generated by GPT-4, ensuring minimal noise in the data annotation pipeline. In total, we annotated $150K$ grounded visual chat data.

In terms of network architecture, we propose connecting the output features of the Language Model (LLM) with a grounding model to handle grounding tasks, relieving the language model from the burden of vision localization tasks. For this purpose, we use the open-set segmentation and detection model OpenSeeD~\cite{zhang2023simple} as the grounding model, enabling both box and pixel-level grounding simultaneously.

To evaluate the capability of grounded visual chat, we introduce the Grounding Bench, a benchmark that assesses grounding and chat performances concurrently. Built upon the foundation of LLaVA bench, our benchmark evaluates chat and phrase grounding in three contexts: conversation, detailed description, and complex reasoning. Additionally, recognizing that grounded detailed description is the most challenging aspect of grounded visual chat, we propose grounded recall and precision metrics. Grounded recall measures the proportion of ground-truth instances correctly mentioned and grounded, while grounded precision measures the accuracy of groundings or predicted boxes. We also calculate the $F_1$ score, a combination of precision and recall. To evaluate the correctness of semantic matching since the models generate free-form phrases, we rely on GPT-4.
\\
\begin{table}
  \centering
  \footnotesize
    \resizebox{0.9\linewidth}{!}{
    \begin{tabular}{lllll|llll}
    \toprule
             & \multicolumn{4}{c|}{input} & \multicolumn{4}{c}{output} \\
             & text & click & box & mark & text  & box & mask & mark \\
             \midrule
    LLaVA~\cite{li2023llava}    &   \checkmark   &       &     &      &       \checkmark&       &     &      \\
    MiniGPT-4~\cite{zhu2023minigpt} &  \checkmark    &       &     &      &       \checkmark&       & &      \\
    GPT4ROI~\cite{zhang2023gpt4roi}  &  \checkmark    &       &  \checkmark   &      &       \checkmark&       &     &      \\
    Shikra~\cite{chen2023shikra}   &    \checkmark  &       &     &      &       \checkmark&       &     &      \\
    Ferret~\cite{you2023ferret}   &   \checkmark   &    \checkmark   &     &      &       \checkmark&     \checkmark  &     &      \\
    MiniGPTv2~\cite{chen2023minigptv2}  &  \checkmark    &       &     &      &       \checkmark&  \checkmark   &     &      \\
    LLaVA1.5~\cite{liu2023improvedllava}   &  \checkmark    &       &     &      &       \checkmark&  \checkmark   &     &      \\
    CogVLM-Grounding~\cite{wang2023cogvlm}   &  \checkmark    &       &     &      &       \checkmark&  \checkmark   &     &      \\
    \shortname{} (Ours)     &    \checkmark  &    \checkmark   &   \checkmark  &   \checkmark   &  \checkmark     &    \checkmark   &   \checkmark  &     \checkmark  \\
    \bottomrule
    \end{tabular}
    }
  \caption{\footnotesize A comparison of input referring and output grounding format of LMMs.}
  \label{tab:inout_format}
  \vspace{-6mm}
\end{table}
In summary, our contributions are as follows:
\begin{enumerate}
\item We introduce a data annotation pipeline to label high-quality Grounded Visual Chat (GVC) data. Leveraging human-labeled object detection data~\cite{lin2014microsoft} and harnessing the robust matching capability of GPT-4~\cite{openai2023gpt4}, we have successfully labeled 150K GVC instances using the LLaVA instruction tuning dataset.

\item We present an end-to-end model, named \fullname~(\shortname{} for brevity), which connects a Large Multimodal Model (LMM) with a grounding model to facilitate grounded visual chat. Our model supports both object and pixel-level grounding, accommodating various visual prompts such as mark, click, box, and scribble. Table~\ref{tab:inout_format} demonstrates that our model offers a broader range of input and output prompt types compared to other LMMs.

\item We establish the \benchname benchmark for evaluating grounded visual chat and propose an auto-evaluation pipeline aided by GPT-4. This benchmark assesses grounded visual chat capabilities and provides performance metrics for other state-of-the-art methods.

\item Through extensive experiments, we demonstrate that our model surpasses other grounding LMMs in terms of performance on \benchname, while also achieving competitive results on classic grounding benchmarks like RefCOCO/+/g and Flickr30K.
\end{enumerate}
\section{Method}

\subsection{Overview}
\begin{figure}[b!]
\centering  
\vspace{-3mm}
\includegraphics[width=0.4\textwidth]{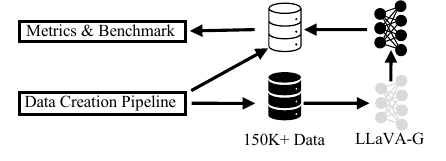} \\
\vspace{-2mm}
\caption{\footnotesize An overview of our main contributions. We use the data creation pipeline to create training and test data. The training data is used to train our \shortname{}. The test data is used to build our \benchname{}.}
\label{fig:overview}  
  \vspace{-3mm}
\end{figure}

To advance the development of grounded visual chat for Large Multimodal Models (LMMs), we introduce a comprehensive pipeline for labeling grounded visual chat data, a tailored modeling approach designed for the grounded visual chat task, and a benchmark for evaluating grounded visual chat performance, as illustrated in Figure~\ref{fig:overview}. We will provide further details on these three components in the following subsections.

\subsection{Grounded Visual Chat Data Creation}
\label{sec: data}
\begin{table*}[t!]\centering
\begin{minipage}{2.0\columnwidth}\vspace{0mm}    \centering
\resizebox{0.8\textwidth}{!}{\begin{tcolorbox} 
\centering
\footnotesize

\begin{tabular}{p{1.0\columnwidth} c}
\VarSty{ {\bf Context type 1: Boxes (for data annotation)} } & \\
1.person: [0.681, 0.242, 0.774, 0.694],
2.person: [0.63, 0.222, 0.686, 0.516],
& \hspace{-6.8cm} 
\multirow{5}{*}
{ \includegraphics[height=3.5cm]{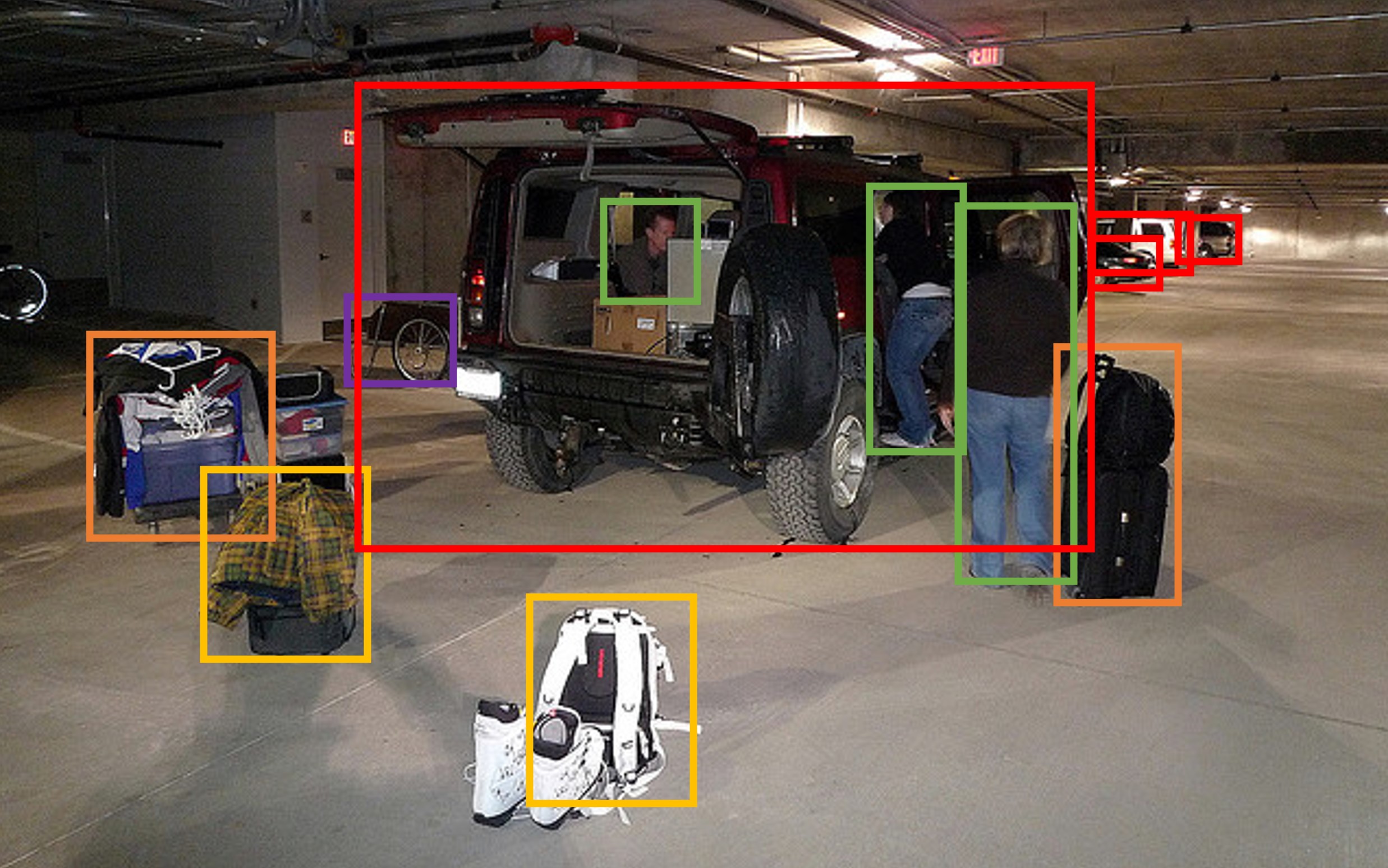} }\\
3.person: [0.444, 0.233, 0.487, 0.34],
4.backpack: [0.384, 0.696, 0.485, 0.914],\\
5.backpack: [0.755, 0.413, 0.846, 0.692],
6.suitcase: [0.758, 0.413, 0.845, 0.69],\\
7.suitcase: [0.1, 0.497, 0.173, 0.579],
8.bicycle: [0.282, 0.363, 0.327, 0.442],\\
9.car: [0.786, 0.25, 0.848, 0.322],
10.car: [0.783, 0.27, 0.827, 0.335],\\
11.car: [0.86, 0.254, 0.891, 0.3],
12.car: [0.261, 0.101, 0.787, 0.626]& \\
\VarSty{ {\bf Context type 2: user responses (for data annotation)} }& \\
The image is an underground parking area with a black sport utility vehicle (SUV)\\ parked. There are three people in the scene, with one person standing closer to the left\\ side of the vehicle, another person in the middle, and the third person on the right side.\\ They are all working together to pack their luggage into the SUV for a trip. \\
\hrulefill & \\
   \VarSty{ {\bf Response: grounded responses (for data annotation)} } & \\
The image is an underground parking area with a (\textcolor{red}{black sport utility vehicle}) [10.car] (SUV) parked. There are (\textcolor{green1}{three people}) [1.person, 2.person, 3.person] in the scene, with (\textcolor{green1}{one person}) [3.person] standing closer to the left side of the vehicle, (\textcolor{green1}{another person}) [2.person] in the middle, and (\textcolor{green1}{the third person}) [1.person] on the right side. They are all working together to pack (\textcolor{orange1}{their luggage}) [4.backpack, 5.backpack, 6.suitcase, 7.suitcase] into the SUV for a trip.
 \end{tabular}
\end{tcolorbox}}

\resizebox{0.8\textwidth}{!}{\begin{tcolorbox} 
\centering
\footnotesize
\begin{tabular}{p{1.0\columnwidth} c}
\PredSty{ {\bf Context type 3: predicted grounded responses (for evaluation)} } & \\
The depiction is of a below-ground parking facility, where \textcolor{red}{a sleek, black vehicle} [9.car] is situated. In the vicinity of this SUV, \textcolor{green1}{a trio of individuals} [1.person, 2.person, 3.person] is engaged in an activity:  \textcolor{green1}{the first person} [1.person] is adjacent to the left side of the vehicle,  \textcolor{green1}{the second} [2.person] is situated centrally, and  \textcolor{green1}{the third} [3.person] is near the right side. They are collaboratively arranging their travel bags in the SUV, signaling the onset of an impending journey.\\
\hrulefill\\
\PredSty{ \bf Response: $TP_{pred}$ and $TP_{gt}$ (for evaluation) }& \\
\textcolor{red}{"a sleek, black vehicle"} [9.car] - Incorrectly referred. \\
\textcolor{green1}{"a trio of individuals"} [1.person, 2.person, 3.person] - 3 Correctly referred. \\
\textcolor{green1}{"the first person"} [1.person] - Incorrectly referred. \\
\textcolor{green1}{"the second"} [2.person] - Correctly referred. \\
\textcolor{green1}{"the third"} [3.person] - Incorrectly referred. \\ \\\textbf{There are 4 correct references ($TP_{pred}$) and 3 correctly referred entities ($TP_{gt}$).}
 \end{tabular}
\end{tcolorbox}}
\vspace{-2mm}
\caption{\footnotesize Illustrate the data annotation (top) and the evaluation (bottom) with language GPT4. The top table shows the contexts and responses for data annotation. The bottom table shows the contexts and responses for evaluating the recall and precision of grounded description. Note that the Context 1 for evaluation is same as that for data annotation. Note that the visual image is not used to prompt GPT4, we only show it here as a reference. }
    \label{tab:full_example_car_bbox}
    \vspace{-0.3cm}
\end{minipage}
\end{table*}
To perform grounded visual chat (GVC) effectively, it is crucial to have high-quality data that encompasses both meaningful conversations and accurate grounding. We have constructed our dataset based on LLaVA instruction tuning data for two primary reasons. Firstly, the conversations within this dataset are generated by GPT-4, known for its high linguistic quality. Secondly, the images used are sourced from COCO, which contains human-annotated grounding box instances.

Our data annotation process aims to associate phrases from conversations with specific instances. To achieve this, we leverage the capabilities of GPT-4. As illustrated in Table~\ref{tab:full_example_car_bbox}, we provide GPT-4 with ground-truth (GT) boxes containing class labels and a sentence from the conversation. We task GPT-4 with matching noun phrases from the sentence to the GT instances. Once noun phrases are successfully grounded by GPT-4, we mark them with special start tokens, $\langle g_s \rangle$ and $\langle g_e \rangle$, followed by a token, $\langle seg \rangle$, which corresponds to the output feature used by the grounding model to segment the grounded region. An example of a question and its answer in the dataset is as follows:

\textit{Q: What is the man doing? A: $\langle g_s \rangle$ The man $\langle g_e \rangle$ $\langle seg \rangle$ is using $\langle g_s \rangle$ a clothing iron $\langle g_e \rangle$ $\langle seg \rangle$ on the back of $\langle g_s \rangle$ a yellow taxi $\langle g_e \rangle$ $\langle seg \rangle$.}

For each $\langle seg \rangle$, we have a corresponding segmentation mask. This annotated data forms the basis of our Grounded Visual Chat (GVC) dataset. Optionally, to support visual prompts in user instructions, we apply a similar annotation process to instances in the question itself. The resulting data appears as follows:

\textit{Q: What is the object $\langle obj \rangle$ doing? A: $\langle g_s \rangle$ The man $\langle g_e \rangle$ $\langle seg \rangle$ is using $\langle g_s \rangle$ a clothing iron $\langle g_e \rangle$ $\langle seg \rangle$ on the back of $\langle g_s \rangle$ a yellow taxi $\langle g_e \rangle$ $\langle seg \rangle$.}

It's important to note that we modify "the man" to "the object" in cases where the model might disregard the visual prompts. For each $\langle obj \rangle$ in the question, we provide a corresponding segmentation mask. This dataset is referred to as GVC-R (Grounded Visual Chat with Referring).
%

\subsection{Network Architectures}
\begin{figure}[t!]
\centering  
\vspace{-0mm}
\includegraphics[width=0.45\textwidth]{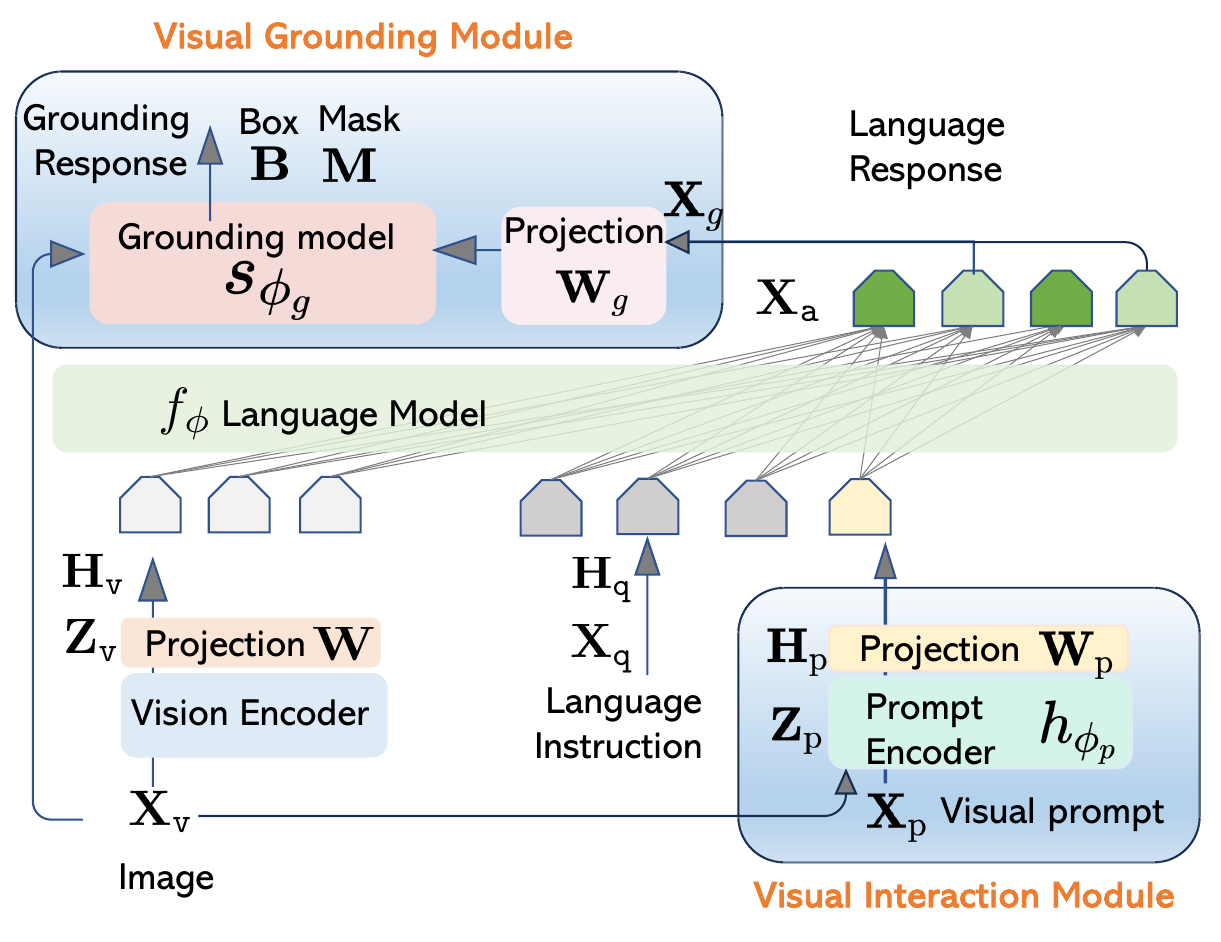} \\
\vspace{-2mm}
\caption{\footnotesize Network architecture of our \fullname{} contains a CLIP vision encoder, a LLM, a prompt encoder, a grounding model and the corresponding projection layers. \fullname{} expands LLaVA with two additional modules highlighted in blue blocks: the visual interaction module that accepts user drawing and visual grounding module that outputs object masks/boxes. The yellow tokens represents the visual prompt feature aligned to language embedding space. The light green output tokens represent the grounding features which are the last-layer hidden feature of the language model corresponding to $\langle seg \rangle$ tokens.}
\label{fig:modeling}  
  \vspace{-3mm}
\end{figure}
Since our network architecture is nearly identical to LLaVA, with the exception of the additional prompt encoder and grounding model, we will only introduce these two parts in this section. For the other components of our architecture, please refer to LLaVA~\cite{liu2023visual}.

\noindent\textbf{Prompt encoder.} For an input image $\Xmat_{\texttt{v}}$ and a visual prompt $\Xmat_{\texttt{p}}$, we employ the pre-trained Semantic-SAM as the prompt encoder. This encoder extracts visual features based on the input image and visual prompts, denoted as $\Zmat_{\texttt{p}}=h(\Xmat_{\texttt{v}},\Xmat_{\texttt{p}})$. To convert these prompt features into language embedding tokens $\Hmat_{\texttt{p}}$ of the same dimensionality as the word embedding space in the language model, we use a simple linear layer with a trainable projection matrix $\Wmat_{\texttt{p}}$:

\begin{equation}
\Hmat_{\texttt{p}}=\Wmat_{\texttt{p}} \cdot \Zmat_{\texttt{p}}, \text{ where } \Zmat_{\texttt{p}}=h\left(\Xmat_{\texttt{v}},\Xmat_{\texttt{p}}\right)
\end{equation}

This results in a sequence of visual tokens $\Hmat_{\texttt{p}}$. It's worth noting that there are special tokens $\langle obj \rangle$ in $\Xmat_{\texttt{q}}$ with word embeddings as placeholders, and visual tokens in $\Hmat_{\texttt{p}}$ replace the word embeddings of $\langle obj \rangle$ in $\Hmat_{\texttt{q}}$.

\noindent\textbf{Grounding model.} In addition to the language response $\Xmat_{\texttt{a}}$, our model also produces features $\Xmat_{\texttt{g}}$ for grounding. These features correspond to the last layer hidden features of the language model that align with the $\langle seg \rangle$ tokens. We initially map these features to a grounding space using a trainable projection matrix $\Wmat_{\texttt{g}}$. Subsequently, we employ a pretrained OpenSeeD model as the grounding model to generate bounding boxes $\mathbf{B}$ and masks $\mathbf{M}$. This process can be defined as follows:

\begin{equation}
\mathbf{B, M}=s\left(\Xmat_{\texttt{v}},\Wmat_{\texttt{g}} \cdot \Xmat_{\texttt{g}}\right)
\end{equation}

Here, $s(\cdot,\cdot)$ represents the grounding model, which takes the image $\Xmat_{\texttt{v}}$ and the grounding features as input.

\subsection{Training}
We propose a three-stage training strategy, as illustrated in Table~\ref{tab: data tasks}. These stages are \textbf{pretraining for alignment}, \textbf{instruction
tuning for grounded visual chat}, and \textbf{extension to visual prompt}. A unified representation of our instruction-following data is presented as follows:
\begin{equation}
\raggedright
    \small
\begin{aligned}
    &\texttt{Human}: \Xmat_{\texttt{v}} ~< \textbackslash \texttt{n}>~\Xmat_{\texttt{q}} (\Xmat_{\texttt{p}})     \color{mygreen!90!black}{\texttt{<STOP>}} \\
    &\texttt{Assistant}: 
    \color{mygreen!90!black}{\Xmat_{\texttt{a} } (\Xmat_{\texttt{g}})\texttt{<STOP>}} \textbackslash \texttt{n}
\end{aligned}
 \end{equation}
 In this representation, $\Xmat_{\texttt{p}}$ and $\Xmat_{\texttt{g}}$ are enclosed in brackets, indicating that they are optional. During training, the model is trained to predict the assistant's answers, including the grounded instances and where to stop.  Consequently, only the {\color{mygreen} green sequence/tokens} are used to compute the loss in the auto-regressive model.
 
\paragraph{Stage 1: Pretraining for alignment.}
Stage 1 focuses on feature alignment for the visual encoder and granularity alignment for the grounding model.\\
\noindent\textbf{Feature alignment for vision encoder.} As shown in Table~\ref{tab: data tasks}, we utilize the RefCOCO/+/g, COCO 2017train, Visual Genome, LLaVA 585K image caption, and Flickr30K Entities datasets for Stage 1. Both LLaVA 585K and Flickr30K Entities datasets consist of image caption pairs and are used to train the projection layer $\mathbf{W}$ for feature alignment in the vision encoder. The conversation construction approach aligns with that of LLaVA, where a question is randomly selected from Table~\ref{tab:short caption} as $\Xmat_{\texttt{q}}$, and the original caption is used as $\Xmat_{\texttt{a}}$. The learnable parameter for this part is denoted as $\theta=\left\{\mathbf{W}\right\}$.

\noindent\textbf{Feature and granularity alignment for grounding model.} To facilitate grounding, we need to align the features $\Xmat_{\texttt{g}}$ output by the language model with the vocabulary space of the grounding model. For this purpose, we train on the RefCOCO/+/g, COCO 2017train, Visual Genome, and Flickr30K Entities datasets. The approach to construct instruction-following data is as follows:
\begin{enumerate}
    \item For RefCOCO/+/g and Visual Genome, the user instruction $\Xmat_{\texttt{q}}$ is randomly selected from Table~\ref{tab:res}, and $\Xmat_{\texttt{a}}$ consists only of the special token $\langle seg \rangle$. COCO 2017train follows the same approach as RefCOCO/+/g, but with a distinction: the class name of an instance serves as its referring text.
    \item In contrast, the Flickr30K Entities dataset differs from the image caption data mentioned earlier. Here, the user instruction is followed by a suffix randomly chosen from Table~\ref{tab:grounded caption}. This suffix signals the model to produce a response in grounding format, as described in Section~\ref{sec: data}. The response $\Xmat_{\texttt{a}}$ is then converted into the grounding format by inserting special tokens $\langle g_s \rangle$, $\langle g_e \rangle$, and $\langle seg \rangle$ into $\Xmat_{\texttt{a}}$ to mark noun phrases.
\end{enumerate}
Given the instruction-following data, the last-layer hidden features of the language model corresponding to $\langle seg \rangle$ tokens $\Xmat_{\texttt{g}}$ are mapped to the grounding vocabulary space by multiplying them with $\Wmat_{\texttt{g}}$. Additionally, since our grounding model is pretrained on COCO and Object365, which have different granularities compared to the Visual Genome and Flickr30K grounding data, we also train the grounding model to align these granularities.

In summary, the learnable parameters for Stage 1 are denoted as $\theta=\left\{\Wmat, \Wmat_{\texttt{g}}, \phi_{g} \right\}$.\\
\begin{table}[]
\resizebox{0.45\textwidth}{!}{
\begin{tabular}{lcccc}
\toprule
              & Grounding & Grounding Seg& Visual Chat& Chat with VP \\
              \midrule
RefCOCO/+/g~\cite{yu2016modeling,kazemzadeh2014referitgame}  &   \VarSty{\checkmark}&    \VarSty{\checkmark}    &               &      \textcolor{red}{\checkmark}         \\
Visual Genome~\cite{krishna2016visual} &     \VarSty{\checkmark}      &  &              &    \textcolor{red}{\checkmark}            \\
COCO train2017~\cite{lin2014microsoft} &     \VarSty{\checkmark}      &  \VarSty{\checkmark} &              &               \\
LLaVA 585K~\cite{liu2023visual}    &           &    &    \VarSty{\checkmark}       &                \\
Flickr30K ~\cite{plummer2015flickr30k}    &    \VarSty{\checkmark}       &  \VarSty{\checkmark}    &   \VarSty{\checkmark}      & \\
LLaVA 150K~\cite{liu2023visual}&           &     &   \textcolor{green1}{\checkmark}
& \\
GVC~\ref{sec: data}&    \textcolor{green1}{\checkmark}       &  \textcolor{green1}{\checkmark}    &   \textcolor{green1}{\checkmark}      & \\
GVC-R~\ref{sec: data}&           &      &         &\textcolor{red}{\checkmark} \\
\bottomrule \\           
\end{tabular}
}
\vspace{-5pt}
\caption{\footnotesize Blue, green and red means the training data and tasks in the 1st, 2nd, and 3rd stages, respectively. "Grounding" means only predict boxes and "Grounding Seg" means predict masks. For Flickr30K, we use SAM to label pseudo GT masks. ``Chat with VP" means chat with visual prompts.}
\label{tab: data tasks}
\end{table}
\paragraph{Stage 2: Instruction tuning for grounded visual chat.}
In the second training stage, we leverage the Grounded Visual Chat (GVC) data, excluding visual prompts, for instruction tuning. To also support chat without grounding, we incorporate LLaVA 158K instruction-following data. During this stage, we freeze the CLIP vision encoder and focus on fine-tuning the other components of the model. The learnable parameters in this stage are denoted as $\theta=\left\{\Wmat, \Wmat_{\texttt{g}}, \phi, \phi_{g} \right\}$.

The data format consists of instruction data containing $\langle seg \rangle $ tokens in the answer, accompanied by several grounding annotations. The number of grounding annotations corresponds to the number of $\langle seg \rangle$ tokens present. In this stage, we calculate both language loss and grounding losses. The language loss is computed in the same manner as in LLaVA for the answer tokens and "STOP" tokens. The grounding losses encompass box, mask, and matching losses. Box and mask losses are utilized solely for training the grounding model, while the matching loss is propagated to the language model.

\paragraph{Stage 3: Extension to visual prompt.} In the third stage, we introduce support for visual prompts as an additional component by training only $h_{\phi_p}$ and the projection layer $\Wmat_{\texttt{p}}$. As detailed in Table~\ref{tab: data tasks}, the training data includes RefCOCO/+/g, Visual Genome, and GVC-R. In contrast to Stage 1, for RefCOCO/+/g and Visual Genome, we provide visual prompts for the ground truth (GT) instances and instruct the model to predict captions. The text instruction $\Xmat_{\texttt{p}}$ is randomly selected from Table~\ref{tab:region caption}, where $\langle obj \rangle$ tokens serve as placeholders, and their input embeddings will be replaced by prompt features. The text answer $\Xmat_{\texttt{a}}$ comprises the original referring expressions.

In this stage, the learnable parameters are represented as $\theta=\left\{\phi_p, \Wmat_{\texttt{p}}\right\}$, where $\phi_p$ is trained to output boxes and masks corresponding to visual prompts, and $\Wmat_{\texttt{p}}$ is trained to align visual prompt features with the language embedding space.

\paragraph{Set-of-Mark (SoM) prompts. (Optional)}
In addition to visual prompts (such as clicks and boxes) that can be handled through the prompt encoder, our model also supports marks as visual prompts, similar to the approach presented in~\cite{yang2023setofmark}. These marks consist of alphanumerics and masks that are directly overlaid on the image. To illustrate, consider the data sample in Sec.\ref{sec: data}. Let's assume we overlay marks labeled as $\langle 1\rangle$, $\langle 2\rangle$, and $\langle 3\rangle$ on the "man," "iron," and "taxi" in the input image. This results in the Grounded and Referring Visual Chat (GRVC) data taking the form:

\textit{Q: What is the object $\langle 1\rangle$ doing? A: The man $\langle 1\rangle$ is using a clothing iron $\langle 2\rangle$ on the back of a yellow taxi $\langle 3\rangle$.}

It's important to note that both the question and answer consist of text only. Therefore, in order to support marks as visual prompts, we specifically fine-tune the language part of the model.
\subsection{\benchname{}}
\paragraph{Benchmark Creation.} We introduce a benchmark named \benchname{} to assess a model's grounded visual chat capability. To evaluate both grounding and chat abilities concurrently, we build this benchmark on top of LLaVA Bench (COCO), which comprises chat data generated by GPT4 and instance annotations from MSCOCO. To enhance the robustness of \benchname{}, we expand our test dataset to include 1000 images with 7000 entities, all sourced from the MSCOCO 2014val split. These images are converted into grounded visual chat data using our data creation pipeline, forming the basis of our test dataset.

\paragraph{Task Definition.} Grounded visual chat tasks involve taking an image $X_{V}$ and a user instruction $I$ as input and generating a caption $T$ accompanied by bounding boxes $\mathbf{b}$, with each bounding box corresponding to a specific phrase.

\paragraph{Evaluate Chat Scores.} Our benchmark evaluation encompasses two main aspects: chat scores and grounded response scores. We outline the evaluation process for \benchname{} in Algorithm 1. Chat scores are akin to those used in LLaVA Bench. However, in contrast, we instruct the model to produce grounded responses. Subsequently, we process the output to remove special tokens and boxes, yielding the pure-text response for evaluation.

\paragraph{Evaluate Grounded Response Scores.} For grounded responses, we specifically evaluate the grounded detailed description task. Our evaluation includes metrics such as recall ($R$) for completeness, precision ($P$) for hallucination, and the $F_1$ score ($F_1$) to combine both aspects. $R$ measures the proportion of entities correctly mentioned and grounded in the description, while $P$ assesses the proportion of correctly predicted groundings. A grounding is deemed correct only when the box matches a ground truth (GT) box with an IoU greater than 0.5, and their semantics are accurately matched. To determine $TP_{pred}$ and $TP_{gt}$ for GPT4, we provide Context types 1 and 3, as shown in the bottom block in Table~\ref{tab:full_example_car_bbox}. For example, in the provided example, $N_{pred}=7$ and $N_{gt}=12$. Based on GPT4's response, we calculate $TP_{pred}=4$ and $TP_{gt}=3$. Consequently, we obtain $P=0.57$, $R=0.25$, and $F_1=0.35$.
\begin{figure}[t]
\begin{minipage}{1.0\linewidth}
  \centering
  \includegraphics[width=1.0\linewidth]{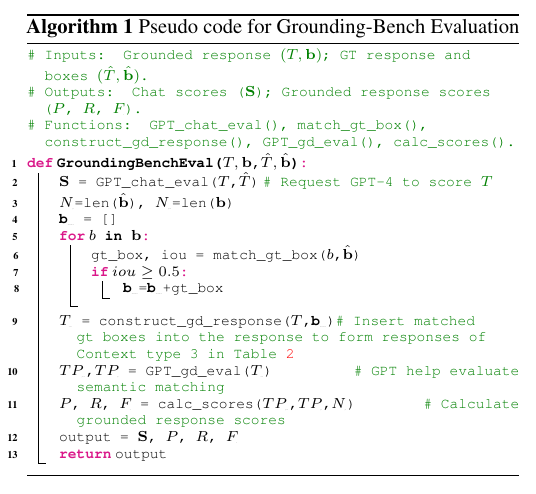}

  \label{fig: algo}
\end{minipage}
\vspace{-4mm}
\end{figure}

\begin{table*}[t!]
\centering
\scalebox{0.7}{
\begin{tabular}{l|c|ccc|cccc|c}
\toprule
  &\#Vision& \multicolumn{3}{c|}{Grounded Response Scores}& \multicolumn{4}{c|}{Chat Scores}&Phrase\\


Model& params(B)&Recall&Precision&$F_1$&Detail desc.& Conv.& Reasoning & \textbf{All}  &\multicolumn{1}{c}{grounding}\\
\midrule
LLaVA~\cite{liu2023visual} &0.30&- &- &-& 69.1 & 82.0 & 92.6 & 81.2 & -\\
Bubo-GPT~\cite{zhao2023bubogpt}       &2.00& {$26.2|25.7$}& {$37.2|31.3$}& {$30.7|28.2$}& {65.0}   & {75.9}   & {93.4}    & {78.2}& -\\
Shikra~\cite{chen2023shikra}    &0.30& $21.1|21.6$& $39.8|38.4$& $27.6|27.7$& 64.7 & 75.4 & 86.4 & 75.5 & 64.29\\
Shikra$^*$          &0.30& $22.0|28.7$& $44.6|48.6$& $29.4|36.1$& 41.8 & -    & -    & -    & - \\
miniGPT v2~\cite{chen2023minigptv2} &1.00& $20.6|25.3$& $33.6|39.1$ & $25.6|30.7$ & 48.0& 51.0& 38.7& 45.8& -\\
CogVLM-Grounding~\cite{wang2023cogvlm}     &10.0& $22.3|27.5$& $56.3|62.5$& $32.0|38.2$ & 35.8& 47.8& 22.2& 34.9& - \\
CogVLM-Chat          &10.0& -& -& -&  73.1& 86.9& 92.1& 84.2& - \\
\textcolor[rgb]{0.753,0.753,0.753}{GPT4-V+SoM~\cite{gpt4v,yang2023setofmark}}      &\textcolor[rgb]{0.753,0.753,0.753}-& \textcolor[rgb]{0.753,0.753,0.753}{$--|55.1$}&\textcolor[rgb]{0.753,0.753,0.753}{$--|73.5$}& \textcolor[rgb]{0.753,0.753,0.753}{$--|63.2$}& \textcolor[rgb]{0.753,0.753,0.753}{67.3}    & \textcolor[rgb]{0.753,0.753,0.753}{104.3}    & \textcolor[rgb]{0.753,0.753,0.753}{108.4}    & \textcolor[rgb]{0.753,0.753,0.753}{93.3}&\textcolor[rgb]{0.753,0.753,0.753}- \\

\midrule
\rowcolor{gray!15} \shortname{} (Ours)     & 0.35&      $28.6|36.3$&   $52.7|53.4$&   $37.1|43.2$& 67.2  & 78.7  & 91.1  &  79.3 &81.6   \\
\bottomrule
\end{tabular}
}

\vspace{-5pt}
\caption{A comparison on our \benchname{}.  For each model, we use the prompt template recommended by the paper. The results in grounded response scores are two parts in each grid where the left one is evaluated on the $1000$ images of our \benchname{} and the right one is on the $30$ images in LLaVA Bench (COCO). $^*$ denotes Shikra with a special prompt for grounded description recommended by the paper. We make GPT4-V+SoM grey because it uses external model to label marks.}
\label{tab: our bench new}
\vspace{-0.4cm}
\end{table*}
\section{Experiments}
In this section, we will first introduce our experimental settings. Then, we will compare our model with other state-of-the-art models on our benchmark, \benchname{}. Next, we will evaluate our model against other grounding models on challenging Referring Expression Comprehension (REC) and Referring Expression Segmentation (RES) tasks on RefCOCO, RefCOCO+, and RefCOCOg. The results will demonstrate that our model outperforms other grounding LLMs with the same number of parameters on both REC and RES tasks, and ours is the only model capable of handling both REC and RES effectively. Afterward, we will conduct an evaluation of the support for various types of visual prompts. Finally, we will perform ablation studies on our modeling and data creation processes to validate our method.
\subsection{Experimental Settings}
To facilitate result reproduction, we provide detailed settings. Our language model is initialized from a pretrained Vicuna-7b v1.3, the grounding model is initialized from the vision part of an OpenSeeD Tiny model pretrained on COCO and Object365, and the interactive encoder is initialized from a Semantic-SAM Tiny model pretrained on COCO with three granularities.

In the first training stage, we freeze the language model and train the grounding model, prompt encoder, and projection layers with a learning rate of $1\times 10^{-4}$. For the second stage, we train the language model and projection layers with a learning rate of $2\times 10^{-5}$, while training the grounding model with a learning rate of $1\times 10^{-4}$ while freezing the CLIP vision encoder and the prompt encoder.

\subsection{\benchname{}}
To demonstrate the effectiveness of our method in Grounded Visual Chat (GVC), we compare our method with other strong LMMs that support visual grounding on our benchmark. As shown in Table~\ref{tab: our bench new}, the results in grounded response scores are presented in two parts for each grid. The left one is evaluated on the 1000 images of our \benchname{}, and the right one is on the 30 images in LLaVA Bench (COCO). All the numbers for grounding LMMs are evaluated using their official prompt templates for grounding to ensure the best performance. The results show that our method outperforms all open-source methods in both grounded response scores and chat scores on grounded responses, except for CogVLM-Chat and LLaVA, which are chat models. GPT4-V achieves the best performance on grounded detailed description with the help of SoM, but it is a combination of two models. Among open-source methods, GogVLM is second only to ours in terms of the $F_1$ score for grounded detailed description, but it has the lowest GPT evaluated scores. Shikra's chat scores are second only to ours. We also annotated 30 images in LLaVA Bench (COCO) as grounded detailed description and reported phrase grounding performance of our model and Shikra for reference.

\subsection{Traditional Grounding Benchmarks}
We also evaluate our model on classic grounding benchmarks, including RefCOCO/+/g for Referring Expression Comprehension (REC) and Referring Expression Segmentation (RES), and Flickr30K Entities for Phrase Grounding. For this experiment, we use the 7B language model with the grounding model using the Swin-Tiny backbone. Our model is trained for the first stage with RefCOCO/+/g, Visual Genome, and Flickr30K Entities. Our model stands out as the only LMM that can excel in both REC and RES tasks. On the REC task, our model outperforms all LMMs, except for CogVLM-Grounding, which utilizes a 4B vision model and a 6B connection module. On RES and Phrase grounding tasks, our model surpasses all LMMs. One advantage of our model is its ability to be trained on both box and mask data, allowing us to leverage Visual Genome to enhance our RES performance.
\begin{table*}[t]
\begin{center}
\setlength{\tabcolsep}{3.5pt}
\resizebox{0.75\textwidth}{!}{
\begin{tabular}{l|ccc|ccc|ccc||cc}
\toprule
\multirow{2}{*}{Models} & \multicolumn{3}{c|}{RefCOCO} & \multicolumn{3}{c|}{RefCOCO+} & \multicolumn{3}{c||}{RefCOCOg}& \multicolumn{2}{c}{Flickr30k Entities}\\
 & REC& \multicolumn{2}{c|}{RES}& REC& \multicolumn{2}{c|}{RES}& REC&  \multicolumn{2}{c||}{RES}& &\\
&ACC@0.5&mIoU&cIoU&ACC@0.5&mIoU&cIoU&ACC@0.5&mIoU&cIoU& val&test \\
\midrule
ReLA~\cite{liu2023gres} & -- & -- & 73.80 & -- & -- & 66.00 & -- & -- & 65.00 & -- & -- \\
PolyFormer-L\cite{liu2023polyformer}& --& 76.94& 75.96& --& 72.15& \textbf{69.33}& --& 71.15&69.20& -- & --\\
UniTAB~\citep{yang2022unitab} & 86.32 & -- & -- & 78.70 & -- &  -- & 79.96 & --  &--& 78.76 & 79.58\\
MDETR~\citep{kamath2021mdetr} & 86.75 & -- & -- & 79.52 & -- & -- & 81.64 & --  &--& 82.3 & 83.8\\
GLIP-T$^*$~\cite{li2022grounded} & 50.42 & -- & -- & 49.50 & -- & -- & 66.09 & --  &-- & -- & --\\
GDINO-T~\cite{liu2023grounding} & \textbf{89.19}& --  & -- &81.09 & -- & -- & 84.15 & --  & --& -- & --\\
\midrule
 Kosmos-2$^*$~\cite{peng2023kosmos}&52.32 &-- & --& 45.48&-- &-- & 60.57& --& --& 77.80&78.70\\
LISA-7B~\citep{chen2023shikra} &--&--&74.9&--&--&65.1&--&--&67.9&--&--\\
MiniGPT v2-7B~\citep{chen2023shikra} & 88.06& --& --& 79.58& --& --& 84.19& --&--& -- & -- \\
Shikra-7B~\citep{chen2023shikra} & 87.01 & -- & -- & 81.60 & -- & -- & 82.27 & --  &--& 75.84 & 76.54 \\
Ferret-7B~\cite{you2023ferret} & 87.49 & -- & -- & 80.78 & -- & -- & 83.93 & --  &--& 80.39 & 82.21 \\
\textcolor[rgb]{0.753,0.753,0.753}{CogVLM-Grounding-17B~\cite{wang2023cogvlm}} & \textcolor[rgb]{0.753,0.753,0.753}{93.40} & \textcolor[rgb]{0.753,0.753,0.753}{--} & \textcolor[rgb]{0.753,0.753,0.753}{--} & \textcolor[rgb]{0.753,0.753,0.753}{87.76} & \textcolor[rgb]{0.753,0.753,0.753}{--} & \textcolor[rgb]{0.753,0.753,0.753}{--} & \textcolor[rgb]{0.753,0.753,0.753}{93.02} & \textcolor[rgb]{0.753,0.753,0.753}{--}  &\textcolor[rgb]{0.753,0.753,0.753}{--}& \textcolor[rgb]{0.753,0.753,0.753}{--} & \textcolor[rgb]{0.753,0.753,0.753}{--} \\
\midrule
\rowcolor{gray!15} \shortname{-7B} (Ours) & 89.16 & \textbf{79.68} & \textbf{77.13} & \textbf{81.68} & \textbf{72.92} & 68.79 & \textbf{84.82} & \textbf{74.39}  & \textbf{71.54} & \textbf{83.03} & \textbf{83.62} \\
\bottomrule
\end{tabular}
}
\label{tab:flickr_refcoco}
\end{center}
\vspace{-3mm}
\caption{\footnotesize Performance comparison on the referring expression comprehension (REC) referring expression segmentation (RES) and phrase grounding tasks. We mark the best results with bold. $^*$ denotes the zero-shot results are reported. Since CogVLM-Grounding is a larger model with 4B vision model and 6B connection module, we make it grey.}
\end{table*}

\subsection{Visual Prompts}
\begin{table}[]
\centering
\resizebox{0.4\textwidth}{!}{
\begin{tabular}{l|ccccc}
\toprule
 Model&Ground type &$\alpha$ Mark   &Size Mark     & {val} & { test} \\
\midrule
 Ours&- &-&-& 83.0                            & 83.6     \\
\midrule
 Ours&Mark &  0.4 / 0.4 & 20 & 72.1                             & 73.7                              \\
 Ours&Mark & 0.4 / 0.2 & 30 & 75.1                             & 75.4                              \\
 Ours&Mark & 0.2 / 0.2 &  30 & 76.6                             & 77.9                             \\ \bottomrule
\end{tabular}}
\caption{\footnotesize The top1 accuracy of phrase grounding on Flickr30K. The first row is our original pipeline with grounding model to do phrase grounding. }
\label{tab: mark}
\vspace{-0.2cm}
\end{table}
\begin{table}
\centering
\resizebox{0.45\textwidth}{!}{\begin{tabular}{c c c c c c} 
\toprule
LLava & Shikra & GPT4ROI & PVIT & Ours-T click & Ours-T box \\
\midrule
40 & 53.9 & 64 & 64.5 & 70.8 & 71.5 \\
\bottomrule
\end{tabular}
}
\caption{\footnotesize The comparison on COCO object classification accuracy. Numbers except for our method are from PVIT~\cite{chen2023positionenhanced} paper. They evaluate llava by cropping the regions of GT boxes.}
\label{tab:clickbox}
\vspace{-0.1cm}
\end{table}
We demonstrate our support for various types of visual prompts, including marks, clicks, and boxes. \\
\noindent\textbf{Support for marks as visual prompts.} In Table~\ref{tab: mark}, we showcase our phrase grounding performance on Flikr30K when using marks as visual prompts. It's important to note that for this experiment, we overlay GT masks and marks onto the input image, positioning the marks at the centers of the masks. The table presents results for different $\alpha$ values of masks and sizes of marks. We also include the performance of Shikra-7B and Ours-7B with the grounding model for comparison. The results clearly indicate that our model effectively supports the use of marks for grounding.\\
\noindent\textbf{Support for clicks and boxes as visual prompts.} To further demonstrate our support for visual prompts, we compare object classification accuracy on COCO val2017 in Table~\ref{tab:clickbox}. We report our performances achieved using clicks and boxes as visual prompts. Additionally, we include numbers from the PVIT paper for reference. The results illustrate that our referring accuracy surpasses that of previous works, with boxes yielding better accuracy than clicks.
\begin{table*}
  \begin{minipage}{0.99\textwidth}
\centering  
\vspace{-4mm}
\scalebox{0.88}{
\begin{tabular}{l p{10.9cm} r }
\toprule
 \multicolumn{3}{l}{\bf Grounded description example 1:}  \\
\midrule
User & Describe the following image with grounding. & 
\multirow{2}{*}{\includegraphics[height=4.0cm]{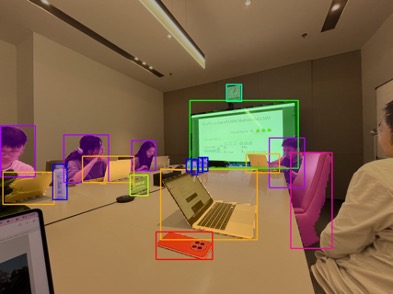}}
\\
\shortname{} & The image depicts \textcolor{people}{a group of people} sitting at a large, white table, engaged in activities. There are four people in the scene, with three of them sitting close to each other and one person sitting further to the right. They are all using laptop computers, likely working or collaborating on a project together. The table is well organized, featuring \textcolor{laptop}{multiple laptops}, \textcolor{TV}{a large TV screen} on the left side, and \textcolor{clock}{a clock} on the wall. There are also \textcolor{books}{several books} and \textcolor{cell_phone}{a cell phone} placed on the table, indicating a casual and comfortable environment. \textcolor{chair}{A chair} is positioned near the left side of the table, and there are \textcolor{bottle}{two bottles} placed nearby, possibly containing drinks for the people working at the table. 
&  \\
\midrule
User & Can you provide a description of the image and include the coordinates [[x0,y0,x1,y1]] for each mentioned object?  & 
\multirow{2}{*}{ \includegraphics[height=4.0cm]{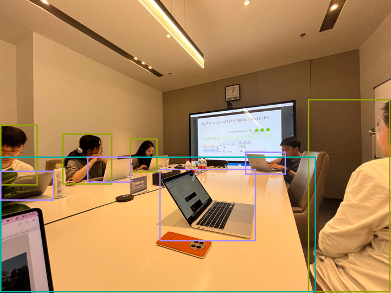}  }
\\
CogVLM-Grounding &  A group of \textcolor{color_182125}{people} are sitting at a \textcolor{color_35381}{table} with \textcolor{color_193416}{laptops}.  \vspace{2.8cm}
& \\
 \midrule
User & Provide a comprehensive description of the image <image> and specify the positions of any mentioned objects in square brackets. 
&
\multirow{2}{*}{\includegraphics[height=4.0cm]{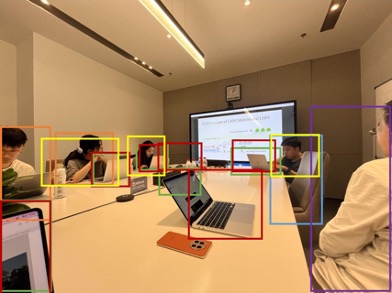} }
 \\
Shikra & A group of \textcolor{color_261472}{people} are in a meeting , \textcolor{color_122260}{some} with \textcolor{color_142857}{laptops}, \textcolor{color_139071}{some} with \textcolor{color_214303}{laptops}, and \textcolor{color_282751}{some} without .  
& \vspace{2.4cm} \\
\midrule
User & [grounding] Describe the scene in detail.  & 
\multirow{2}{*}{ \includegraphics[height=4.0cm]{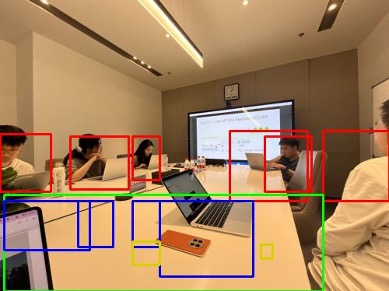} }
\\
MiniGPT v2 & \textcolor{color_274846}{A group of people} are sitting around \textcolor{color_37696}{a large table} in a meeting room , with \textcolor{color_30046}{laptops} and \textcolor{color_238111}{phones} on the table
&  \vspace{2.75cm} \\
\bottomrule
\end{tabular}
}
\captionof{table}{An example of grounded detailed descriptions generated by \shortname{} and other LMMs. We use the best prompt recommended by the paper for each model.}  
\label{tab:visual_example real life}  
  \end{minipage}
\end{table*}
\begin{table*}
  \begin{minipage}{0.99\textwidth}
\centering  
\scalebox{0.88}{
\begin{tabular}{l p{10.9cm} r }
\toprule
 \multicolumn{3}{l}{\bf Visual prompt example:}  \\
\midrule
User & What is the object $\langle obj \rangle$ doing and why?
& 
\multirow{2}{*}{\includegraphics[height=4.0cm]{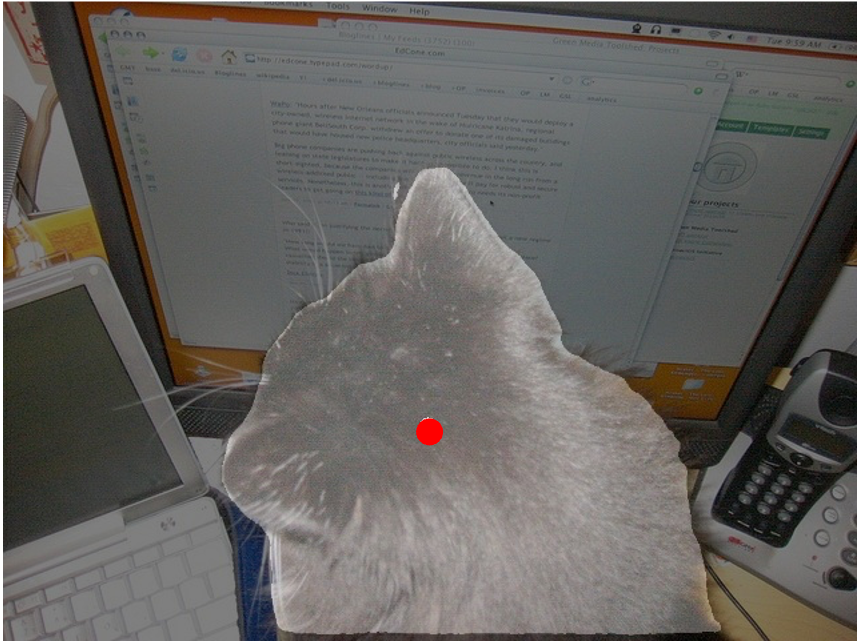}}
\\
\shortname{} & The cat is sitting in front of a laptop computer, likely curious about the content displayed on the screen. Cats are naturally inquisitive animals, and they are often attracted to anything that seems new, interesting, or engaging. The movement or visual elements on the laptop screen might have caught the cat's attention, prompting it to sit and observe the screen. It could be merely a passing curiosity or the start of an interest in technology or screen time that the cat may develop further in the future.
& \vspace{0.8cm} \\
\midrule
User & What is the object $\langle obj \rangle$ used for?
  & 
\multirow{2}{*}{ \includegraphics[height=4.0cm]{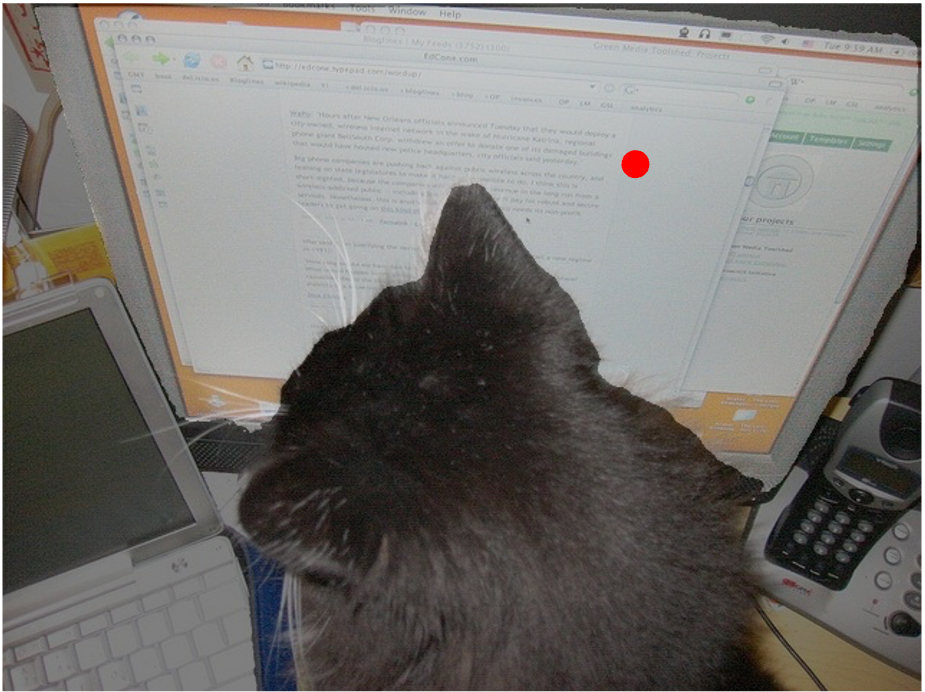} }
\\
\shortname{} &  The laptop computer screen is used for various purposes, such as working, studying, entertainment, and communication. In this particular image, the laptop appears to have a Word document or a web page open, which suggests that the user might be working on a document, browsing the internet, or using the computer for other tasks.
 \vspace{1.7cm}
& \\
\bottomrule
\end{tabular}
}
\captionof{table}{An example of  visual chat based on visual prompts where $\langle obj \rangle$ denotes the positions in the user instruction to insert the visual prompt features.}  
\label{tab:visual prompt example}  
  \end{minipage}
\end{table*}
\begin{table*}
  \begin{minipage}{0.99\textwidth}
\centering  
\scalebox{0.88}{
\begin{tabular}{l p{10.9cm} r }
\toprule
 \multicolumn{3}{l}{\bf Example of marks as prompts:}  \\
\midrule
User & What do you think is going on in this snapshot with grounding?

& 
\multirow{2}{*}{\includegraphics[height=4.0cm]{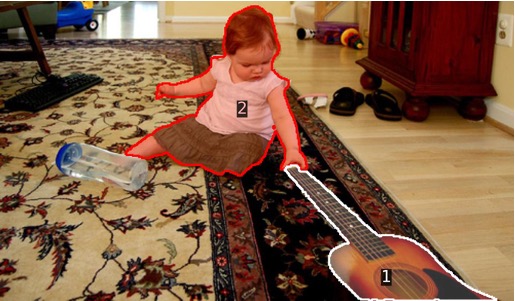}}
\\
\shortname{} & In the snapshot, a little girl <2> is sitting on a rug with a guitar-shaped toy <1> in her hands. She appears to be playing or interacting with the toy guitar, which suggests that she might be interested in music or is simply having fun with her toy. This scene could indicate that the child is exposed to musical instruments or enjoys imitating the sounds she hears. It is also possible that the toy guitar is a part of her imaginative play, where she creates different scenarios and engages with her surroundings.
& \vspace{0.5cm} \\
\bottomrule
\end{tabular}
}
\captionof{table}{An example of  visual chat based on marks.}  
\label{tab:mark example}  
  \end{minipage}
\end{table*}
\begin{table*}
  \begin{minipage}{0.99\textwidth}
\centering  
\scalebox{0.88}{
\begin{tabular}{l p{10.9cm} r }
\toprule
 \multicolumn{3}{l}{\bf Example of reduce hallucination:}  \\
\midrule
User & What do you see happening in this image with grounding?& 
\multirow{2}{*}{\includegraphics[height=4.0cm]{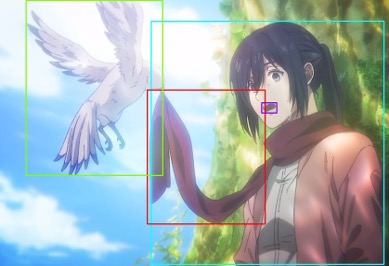}}\\
\shortname{} & The image features \textcolor{color_37951}{a young woman} standing in a field, surrounded by tall grass. She is holding \textcolor{color_274846}{a string or a ribbon} in her hand, which seems to be attached to \textcolor{color_159743}{a bird} . The bird is flying in the sky, with its position slightly to the right and above the woman. In addition to the woman and the bird, there are \textcolor{color_152093}{two small birds} in the scene. One is located in the middle of the scene, close to the woman, while the other is slightly above the first bird. The scene has a peaceful, nature-oriented atmosphere.
& \vspace{0.8cm} \\
\midrule
User & There are actually no small birds, so please describe the scene again with grounding.
  & 
\multirow{2}{*}{ \includegraphics[height=4.0cm]{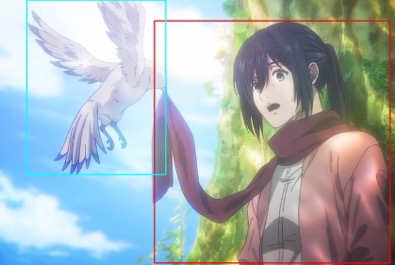} }
\\
\shortname{} & In the image, there is \textcolor{color_274846}{a young woman} standing in a field of tall grass. She holds a string or a ribbon, which is connected to \textcolor{color_37951}{a bird} . The bird is flying in the sky, towards the right and slightly above the woman. The scene portrays a sense of wonder and tranquility as the woman interacts with the bird and enjoys the company of nature.
 \vspace{1.3cm}
& \\
\bottomrule
\end{tabular}
}
\captionof{table}{An example to show the potential of \shortname{} to reduce hallucination.}  
\label{tab:reduce hallucination example}  
  \end{minipage}
\end{table*}
\subsection{Visualizations}
\noindent\textbf{Compare with other models.} We provide visual comparisons between our model and other grounding LMMs, including CogVLM-Grounding, Shikra, and MiniGPTv2, in Table~\ref{tab:visual_example real life}, \ref{tab:visual_example openset}, and \ref{tab:visual_example coco}. These visualizations highlight that our model produces longer descriptions, detects more objects, and maintains high precision. Furthermore, our model demonstrates the ability to ground open-set concepts, such as "dragons," which do not exist in our grounded visual chat data. These visual comparisons underscore the strengths and capabilities of our model in the context of grounded visual chat.\\
\noindent\textbf{Other functionalities.} 
\begin{enumerate}
\item Visual Chat with Visual Prompts (Table~\ref{tab:visual prompt example}): Our model demonstrates its capability to engage in visual chat using visual prompts such as boxes or clicks.
\item Support for Marks as Visual Prompts (Table~\ref{tab:mark example}): After fine-tuning with mark data, our model effectively supports marks as visual prompts.
\item Debugging and Hallucination Removal (Table~\ref{tab:reduce hallucination example}): Our model exhibits a potential for assisting in debugging by allowing users to identify and correct errors. In this example, the initial output contains hallucination ("two small birds" grounded to the woman's mouth), which the user detects through grounding and subsequently instructs the model to rectify.
\end{enumerate}

\subsection{Ablation Studies}
\begin{table}[t!]
    \centering
\centering 
     \resizebox{0.45\textwidth}{!}{
    \begin{tabular}{l|lcccc|cccc}
    \toprule
     &&  \multicolumn{4}{c}{\textbf{LLaVA-Bench (COCO)}} & \multicolumn{4}{c}{\textbf{LLaVA-Bench (In-the-Wild)}} \\
          &GVC&  
        {\footnotesize Conv.} & {\footnotesize Detail} & {\footnotesize Reasoning} & {\footnotesize All}
        & {\footnotesize Conv.} & {\footnotesize Detail} & {\footnotesize Reasoning} & {\footnotesize All}\\ \midrule
        LLaVA    && 82.0  & 69.1  & 92.6  & 81.2  & 42.6  &	51.9  &	68.9 & 57.1  \\
        \midrule
 \shortname{}  &Yes&74.8 &68.5 &95.3 &79.7 &38.5  &40.1  &75.1&55.8   \\ 
        \shortname  &No& 79.3&71.2	&	92.8& 	81.2& 47.7 & 44.6 & 70.0 & 57.2\\
        \bottomrule
    \end{tabular}
    }
\vspace{-5pt}
     \caption{\footnotesize A comparison on LLaVA-Bench. ``GVC" is ``No" means it outputs pure-text response without grounding.}
     \vspace{-1mm}

    \label{tab: language performance}
\end{table}
\begin{table}
\centering
\resizebox{0.45\textwidth}{!}{
\begin{tabular}{l  c c c|c c c|c c c}
\toprule
 & \multicolumn{3}{c|}{RefCOCO} & \multicolumn{3}{c|}{RefCOCO+} & \multicolumn{3}{c}{RefCOCOg}\\
\#Q& ACC& cIoU & mIoU& ACC& cIoU & mIoU&ACC& cIoU & mIoU\\
\midrule
50 & 86.71 & 74.77 & 77.6 &  77.91 & 64.97 & 69.68 &82.37&68.46&72.43\\
100  & 86.58 & 74.70 & 77.40 &  77.23 & 64.08 & 69.02 &81.99&68.02&72.06\\
300  & 86.35 & 74.26 & 77.19 & 77.78 & 64.68 & 69.54 &81.92&67.89&71.85\\
\bottomrule
\end{tabular}}
\vspace{-5pt}
\caption{\footnotesize The comparison of performance when using different number of queries in the grounding model. ``\#Q" denotes the number of queries.}
\vspace{-0.4cm}
\label{tab: query num}
\end{table}
\begin{table}[t!]
\centering
\resizebox{0.5\textwidth}{!}{
\begin{tabular}{l|c|ccc|cccc}
\toprule
 &  Detach& \multicolumn{3}{c|}{Grounded detail description}& \multicolumn{4}{c}{Chat scores}\\


Model   & GD&Recall&Precision&$F_1$&Detail desc.& Conv.& Reasoning & \textbf{All}  \\
\midrule
Ours&\checkmark& $25.1$& $58.2$& $35.1$& 61.6 & 86.3    & 94.9    & 81.2    \\
Ours&& 36.3& $53.4$& $43.2$& 67.2  & 78.7  & 91.1  & 79.3 \\
\bottomrule
\end{tabular}
}
\vspace{-5pt}
\caption{\footnotesize Ablations on our benchmark. ``Detach GD" means stop gradient from the grounding model to language model. }
\label{tab:ablation data detach}
\vspace{-5mm}
\end{table}
In this section, we provide insights into our visual chat capability and the design of the grounding model through various ablation studies.\\
\noindent\textbf{Maintaining visual chat capability.} We demonstrate that our model retains strong visual chat capabilities by comparing it with LLaVA on LLaVA Bench (Table~\ref{tab: language performance}). The results indicate that our model's visual chat performance is comparable to LLaVA, whether responding with or without grounding.\\
\noindent\textbf{Number of queries in grounding model.} Table~\ref{tab: query num} presents our model's performance on Referring Expression Comprehension (REC) and Referring Expression Segmentation (RES) tasks with different numbers of queries. The results reveal that using 50 queries is sufficient for both tasks and achieves optimal performance. This finding highlights the efficiency of our approach in handling these tasks.\\
\noindent\textbf{Detaching the grounding model.} We investigate the impact of detaching the grounding model on both chat and grounding performance. Detaching the grounding model means stopping gradients from propagating from the grounding model to the Language Model (LLM). Table~\ref{tab:ablation data detach} compares the detached model's performance with the original model. The results demonstrate that detaching the grounding model leads to slightly improved chat performance but significantly compromises the grounding performance. This indicates the importance of the grounding model in maintaining high-quality visual chat with grounding capabilities.

\section{Related Work}
\subsection{Large Multi-modal Models}
With the recent surge in Large Language Models (LLMs), researchers have been actively exploring ways to extend the capabilities of these models beyond text to encompass other modalities, resulting in the development of several Large Multi-modal Models (LMMs). Notably, projects like LLaVA~\cite{li2023llava} and MiniGPT-4~\cite{zhu2023minigpt} have undertaken the task of integrating visual instruction tuning data into their LMMs. They achieved this by utilizing GPT-4 or employing hand-designed prompts, thereby enhancing the LMMs' ability to follow instructions effectively.

In addition to these, there exist other noteworthy works in the field, including mPLUG-DocOwl~\cite{ye2023mplug_docowl}, Otter~\cite{li2023otter}, LLaMa-Adaptor~\cite{zhang2023llamaadapter}, and InternGPT~\cite{liu2023interngpt}. These projects have also contributed significantly to the advancement of LMMs by incorporating various techniques and methodologies.

Moreover, researchers have delved into the realm of fine-grained understanding of LMMs, as exemplified by works like VisionLLM~\cite{wang2023visionllm}, GPT4RoI~\cite{zhang2023gpt4roi}, and PVIT~\cite{chen2023positionenhanced}. VisionLLM, for instance, employs a language-guided tokenizer to extract vision features at specific granularities, whereas GPT4RoI and PVIT utilize bounding boxes to obtain relevant visual features.

\subsection{Visual Grounding Models}
The visual grounding task~\cite{kamath2021mdetr, UNINEXT, zhu2022seqtr, deng2022transvg, MCN, liu2023polyformer, luo2023towards} aims to pinpoint the location of objects within an image based on textual input. This challenge is fundamental in multimodal perception and has promising applications. It requires a deep understanding of both the image and the text, along with establishing correspondences between image regions and textual descriptions.

The GLIP model~\cite{glip} takes a significant step in this direction by integrating various data formats, including detection and referring data. It demonstrates that grounded pretraining effectively enhances the localization capabilities of grounding models. Building upon GLIP, GLIPv2~\cite{zhang2022glipv2} takes a further stride by unifying grounding and Visual-Language (VL) understanding tasks. Grounding-DINO~\cite{liu2023grounding}, which leverages grounded pretraining and the DINO~\cite{zhang2022dino} detector, stands out for its superior performance in this domain.

In recent years, vision-and-language models have gained increasing attention in tasks related to visual recognition and perception. Models like CLIP~\cite{CLIP} and ALIGN~\cite{ALIGN}, through contrastive learning on large-scale image-text pair datasets at the image level, have achieved generalized and robust capabilities in image classification. Simultaneously, in more fine-grained recognition tasks like visual grounding~\cite{kamath2021mdetr, UNINEXT, zhu2022seqtr, deng2022transvg, MCN, liu2023polyformer, luo2023towards,huang2022unified}, which aims to locate specific regions based on textual inputs, researchers are exploring the potential of conducting image and text contrastive learning at the region level.

Approaches such as MDETR~\cite{kamath2021mdetr}, DetCLIP~\cite{yao2022detclip}, DetCLIPv2~\cite{yao2023detclipv2}, GLIP~\cite{glip}, GLIPv2~\cite{zhang2022glipv2}, and Grounding-DINO~\cite{liu2023grounding} strive to detect arbitrary categories by training with large-scale region-text data. For instance, MDETR~\cite{kamath2021mdetr} was trained on existing multimodal datasets with explicit alignment between text phrases and image objects, employing an end-to-end framework.

GLIP~\cite{glip} advances this approach by re-formulating object detection as a grounding task and incorporating additional grounding data to perform grounded pretraining, enhancing semantic alignment between phrases and regions. GLIPv2 further demonstrates how grounded pretraining can improve VL understanding, leading to a unified model for localization and VL understanding.

Moreover, Grounding-DINO~\cite{liu2023grounding}, by incorporating grounded pretraining with the DINO~\cite{zhang2022dino} detector, excels in this field. These advancements in vision-and-language models, particularly through contrastive learning on large-scale text-region data, represent significant progress in fine-grained recognition tasks, resulting in more precise and contextually aware visual understanding.

\subsection{Grounding Large Multi-modal Models}
Based on their architectural characteristics and functionalities, Grounding LMMs can be classified into three distinct categories.

The first category involves models that predict box coordinates in text format. Notable models in this category include Kosmos-2~\cite{peng2023kosmos}, Shikra~\cite{chen2023shikra}, MiniGPT v2~\cite{zhu2023minigpt}, Ferret~\cite{you2023ferret}, and CogVLM~\cite{wang2023cogvlm}. For instance, Kosmos-2 introduced a comprehensive grounding caption dataset and trained a model with strong grounding capabilities, showcasing impressive zero-shot performance across various grounding benchmarks. Shikra, on the other hand, focused on building referral dialog data and training their model to support referral dialog with boxes as both input and output. MiniGPT v2 employed task tokens to activate different task-specific capabilities, including support for grounded output with boxes. Meanwhile, CogVLM leveraged a 10-billion parameter vision model to achieve state-of-the-art performance in various vision-language tasks, including grounding. It's worth noting that many of these methods trained on low-quality grounding caption data, despite achieving significant progress in visual grounding. For instance, Shikra's referential dialog data, although valuable, is relatively small, consisting of only 5,000 images.

The second category involves models that employ a separate grounding model for grounded chat, exemplified by BuboGPT~\cite{zhao2023bubogpt} and LLaVA-PLUS~\cite{liu2023llavaplus}. However, these models often face performance limitations at the language encoder of the grounding model.

The third category adopts an approach where the output of a language model is fed into a grounding model to decode masks and boxes. LISA~\cite{lai2023lisa} is a representative model in this category, with a primary focus on various segmentation tasks rather than chat interactions.

In many previous works, there has been a trade-off between grounding and chat abilities, with data and evaluation metrics typically emphasizing one of these aspects. In contrast, our dataset and benchmark prioritize assessing the compositional abilities of both grounding and chat interactions, providing a unique perspective in this field.
\section{Conclusion}
This paper introduced \fullname{}, an AI assistant that combines visual chat and grounding capabilities. We began by creating a grounded visual chat dataset using a novel data creation pipeline. Subsequently, we proposed an end-to-end model architecture that integrates a grounding model with a Language Model (LM) for effective grounding. Additionally, we introduced \benchname{} as a comprehensive benchmark for evaluating grounded visual chat performance, covering both chat and grounding aspects. Our experiments demonstrated that \fullname{} consistently outperforms other open-source LM models in both chat and grounding tasks, showcasing its effectiveness. Furthermore, \fullname{} excelled in traditional grounding benchmarks, highlighting its versatility. However, we acknowledge that \fullname{} has limitations in terms of semantic scope, and future work could explore extending the dataset and data labeling methods to open-vocabulary settings.

{
    \small
    \bibliographystyle{ieeenat_fullname}
    \bibliography{main}

\begin{thebibliography}{49}
\providecommand{\natexlab}[1]{#1}
\providecommand{\url}[1]{\texttt{#1}}
\expandafter\ifx\csname urlstyle\endcsname\relax
  \providecommand{\doi}[1]{doi: #1}\else
  \providecommand{\doi}{doi: \begingroup \urlstyle{rm}\Url}\fi

\bibitem[Chen et~al.(2023{\natexlab{a}})Chen, Qin, Luo, Mi, Li, Sun, and Liu]{chen2023positionenhanced}
Chi Chen, Ruoyu Qin, Fuwen Luo, Xiaoyue Mi, Peng Li, Maosong Sun, and Yang Liu.
\newblock Position-enhanced visual instruction tuning for multimodal large language models, 2023{\natexlab{a}}.

\bibitem[Chen et~al.(2023{\natexlab{b}})Chen, Zhu, Shen, Li, Liu, Zhang, Krishnamoorthi, Chandra, Xiong, and Elhoseiny]{chen2023minigptv2}
Jun Chen, Deyao Zhu, Xiaoqian Shen, Xiang Li, Zechun Liu, Pengchuan Zhang, Raghuraman Krishnamoorthi, Vikas Chandra, Yunyang Xiong, and Mohamed Elhoseiny.
\newblock Minigpt-v2: Large language model as a unified interface for vision-language multi-task learning.
\newblock \emph{arXiv:2310.09478}, 2023{\natexlab{b}}.

\bibitem[Chen et~al.(2023{\natexlab{c}})Chen, Zhang, Zeng, Zhang, Zhu, and Zhao]{chen2023shikra}
Keqin Chen, Zhao Zhang, Weili Zeng, Richong Zhang, Feng Zhu, and Rui Zhao.
\newblock Shikra: Unleashing multimodal llm's referential dialogue magic.
\newblock \emph{arXiv preprint arXiv:2306.15195}, 2023{\natexlab{c}}.

\bibitem[Deng et~al.(2022)Deng, Yang, Chen, Zhou, and Li]{deng2022transvg}
Jiajun Deng, Zhengyuan Yang, Tianlang Chen, Wengang Zhou, and Houqiang Li.
\newblock Transvg: End-to-end visual grounding with transformers, 2022.

\bibitem[Huang et~al.(2022)Huang, Li, Zhang, Liu, Zhang, and Wang]{huang2022unified}
Shijia Huang, Feng Li, Hao Zhang, Shilong Liu, Lei Zhang, and Liwei Wang.
\newblock A unified mutual supervision framework for referring expression segmentation and generation, 2022.

\bibitem[Jia et~al.(2021)Jia, Yang, Xia, Chen, Parekh, Pham, Le, Sung, Li, and Duerig]{ALIGN}
Chao Jia, Yinfei Yang, Ye Xia, Yi-Ting Chen, Zarana Parekh, Hieu Pham, Quoc~V Le, Yunhsuan Sung, Zhen Li, and Tom Duerig.
\newblock Scaling up visual and vision-language representation learning with noisy text supervision.
\newblock \emph{arXiv preprint arXiv:2102.05918}, 2021.

\bibitem[Kamath et~al.(2021)Kamath, Singh, LeCun, Synnaeve, Misra, and Carion]{kamath2021mdetr}
Aishwarya Kamath, Mannat Singh, Yann LeCun, Gabriel Synnaeve, Ishan Misra, and Nicolas Carion.
\newblock Mdetr -- modulated detection for end-to-end multi-modal understanding, 2021.

\bibitem[Kazemzadeh et~al.(2014)Kazemzadeh, Ordonez, Matten, and Berg]{kazemzadeh2014referitgame}
Sahar Kazemzadeh, Vicente Ordonez, Mark Matten, and Tamara Berg.
\newblock Referitgame: Referring to objects in photographs of natural scenes.
\newblock In \emph{Proceedings of the 2014 conference on empirical methods in natural language processing (EMNLP)}, pages 787--798, 2014.

\bibitem[Krishna et~al.(2016)Krishna, Zhu, Groth, Johnson, Hata, Kravitz, Chen, Kalantidis, Li, Shamma, Bernstein, and Li]{krishna2016visual}
Ranjay Krishna, Yuke Zhu, Oliver Groth, Justin Johnson, Kenji Hata, Joshua Kravitz, Stephanie Chen, Yannis Kalantidis, Li-Jia Li, David~A. Shamma, Michael~S. Bernstein, and Fei-Fei Li.
\newblock Visual genome: Connecting language and vision using crowdsourced dense image annotations, 2016.

\bibitem[Lai et~al.(2023)Lai, Tian, Chen, Li, Yuan, Liu, and Jia]{lai2023lisa}
Xin Lai, Zhuotao Tian, Yukang Chen, Yanwei Li, Yuhui Yuan, Shu Liu, and Jiaya Jia.
\newblock Lisa: Reasoning segmentation via large language model.
\newblock \emph{arXiv preprint arXiv:2308.00692}, 2023.

\bibitem[Li et~al.(2023{\natexlab{a}})Li, Zhang, Chen, Wang, Yang, and Liu]{li2023otter}
Bo Li, Yuanhan Zhang, Liangyu Chen, Jinghao Wang, Jingkang Yang, and Ziwei Liu.
\newblock Otter: A multi-modal model with in-context instruction tuning.
\newblock \emph{arXiv preprint arXiv:2305.03726}, 2023{\natexlab{a}}.

\bibitem[Li et~al.(2023{\natexlab{b}})Li, Wong, Zhang, Usuyama, Liu, Yang, Naumann, Poon, and Gao]{li2023llava}
Chunyuan Li, Cliff Wong, Sheng Zhang, Naoto Usuyama, Haotian Liu, Jianwei Yang, Tristan Naumann, Hoifung Poon, and Jianfeng Gao.
\newblock Llava-med: Training a large language-and-vision assistant for biomedicine in one day.
\newblock \emph{arXiv preprint arXiv:2306.00890}, 2023{\natexlab{b}}.

\bibitem[Li* et~al.(2022)Li*, Zhang*, Zhang*, Yang, Li, Zhong, Wang, Yuan, Zhang, Hwang, Chang, and Gao]{glip}
Liunian~Harold Li*, Pengchuan Zhang*, Haotian Zhang*, Jianwei Yang, Chunyuan Li, Yiwu Zhong, Lijuan Wang, Lu Yuan, Lei Zhang, Jenq-Neng Hwang, Kai-Wei Chang, and Jianfeng Gao.
\newblock Grounded language-image pre-training.
\newblock In \emph{CVPR}, 2022.

\bibitem[Li et~al.(2022)Li, Zhang, Zhang, Yang, Li, Zhong, Wang, Yuan, Zhang, Hwang, et~al.]{li2022grounded}
Liunian~Harold Li, Pengchuan Zhang, Haotian Zhang, Jianwei Yang, Chunyuan Li, Yiwu Zhong, Lijuan Wang, Lu Yuan, Lei Zhang, Jenq-Neng Hwang, et~al.
\newblock Grounded language-image pre-training.
\newblock In \emph{CVPR}, 2022.

\bibitem[Lin et~al.(2014)Lin, Maire, Belongie, Hays, Perona, Ramanan, Doll{\'a}r, and Zitnick]{lin2014microsoft}
Tsung-Yi Lin, Michael Maire, Serge Belongie, James Hays, Pietro Perona, Deva Ramanan, Piotr Doll{\'a}r, and C~Lawrence Zitnick.
\newblock Microsoft {COCO}: Common objects in context.
\newblock In \emph{ECCV}, 2014.

\bibitem[Liu et~al.(2023{\natexlab{a}})Liu, Ding, and Jiang]{liu2023gres}
Chang Liu, Henghui Ding, and Xudong Jiang.
\newblock Gres: Generalized referring expression segmentation.
\newblock In \emph{Proceedings of the IEEE/CVF Conference on Computer Vision and Pattern Recognition}, pages 23592--23601, 2023{\natexlab{a}}.

\bibitem[Liu et~al.(2023{\natexlab{b}})Liu, Li, Li, and Lee]{liu2023improvedllava}
Haotian Liu, Chunyuan Li, Yuheng Li, and Yong~Jae Lee.
\newblock Improved baselines with visual instruction tuning, 2023{\natexlab{b}}.

\bibitem[Liu et~al.(2023{\natexlab{c}})Liu, Li, Wu, and Lee]{liu2023visual}
Haotian Liu, Chunyuan Li, Qingyang Wu, and Yong~Jae Lee.
\newblock Visual instruction tuning.
\newblock \emph{arXiv preprint arXiv:2304.08485}, 2023{\natexlab{c}}.

\bibitem[Liu et~al.(2023{\natexlab{d}})Liu, Ding, Cai, Zhang, Satzoda, Mahadevan, and Manmatha]{liu2023polyformer}
Jiang Liu, Hui Ding, Zhaowei Cai, Yuting Zhang, Ravi~Kumar Satzoda, Vijay Mahadevan, and R. Manmatha.
\newblock Polyformer: Referring image segmentation as sequential polygon generation, 2023{\natexlab{d}}.

\bibitem[Liu et~al.(2023{\natexlab{e}})Liu, Cheng, Liu, Zhang, Li, Ren, Zou, Yang, Su, Zhu, Zhang, Gao, and Li]{liu2023llavaplus}
Shilong Liu, Hao Cheng, Haotian Liu, Hao Zhang, Feng Li, Tianhe Ren, Xueyan Zou, Jianwei Yang, Hang Su, Jun Zhu, Lei Zhang, Jianfeng Gao, and Chunyuan Li.
\newblock Llava-plus: Learning to use tools for creating multimodal agents, 2023{\natexlab{e}}.

\bibitem[Liu et~al.(2023{\natexlab{f}})Liu, Zeng, Ren, Li, Zhang, Yang, Li, Yang, Su, Zhu, et~al.]{liu2023grounding}
Shilong Liu, Zhaoyang Zeng, Tianhe Ren, Feng Li, Hao Zhang, Jie Yang, Chunyuan Li, Jianwei Yang, Hang Su, Jun Zhu, et~al.
\newblock Grounding dino: Marrying dino with grounded pre-training for open-set object detection.
\newblock \emph{arXiv preprint arXiv:2303.05499}, 2023{\natexlab{f}}.

\bibitem[Liu et~al.(2023{\natexlab{g}})Liu, He, Wang, Wang, Wang, Chen, Zhang, Lai, Yang, Li, Yu, Li, Chen, Yang, Zhu, Wang, Wang, Luo, Dai, and Qiao]{liu2023interngpt}
Zhaoyang Liu, Yinan He, Wenhai Wang, Weiyun Wang, Yi Wang, Shoufa Chen, Qinglong Zhang, Zeqiang Lai, Yang Yang, Qingyun Li, Jiashuo Yu, Kunchang Li, Zhe Chen, Xue Yang, Xizhou Zhu, Yali Wang, Limin Wang, Ping Luo, Jifeng Dai, and Yu Qiao.
\newblock Interngpt: Solving vision-centric tasks by interacting with chatgpt beyond language, 2023{\natexlab{g}}.

\bibitem[Luo et~al.(2020)Luo, Zhou, Sun, Cao, Wu, Deng, and Ji]{MCN}
Gen Luo, Yiyi Zhou, Xiaoshuai Sun, Liujuan Cao, Chenglin Wu, Cheng Deng, and Rongrong Ji.
\newblock Multi-task collaborative network for joint referring expression comprehension and segmentation.
\newblock In \emph{Proceedings of the IEEE/CVF Conference on Computer Vision and Pattern Recognition (CVPR)}, 2020.

\bibitem[Luo et~al.(2023)Luo, Zhou, Sun, Wu, Gao, and Ji]{luo2023towards}
Gen Luo, Yiyi Zhou, Xiaoshuai Sun, Yongjian Wu, Yue Gao, and Rongrong Ji.
\newblock Towards language-guided visual recognition via dynamic convolutions.
\newblock \emph{International Journal of Computer Vision}, pages 1--19, 2023.

\bibitem[OpenAI(2023{\natexlab{a}})]{gpt4}
OpenAI.
\newblock Gpt-4 technical report, 2023{\natexlab{a}}.

\bibitem[OpenAI(2023{\natexlab{b}})]{gpt4v}
OpenAI.
\newblock Gpt-4v(ision) system card.
\newblock \url{https://cdn.openai.com/papers/GPTV_System_Card.pdf}, 2023{\natexlab{b}}.

\bibitem[OpenAI(2023{\natexlab{c}})]{openai2023gpt4}
OpenAI.
\newblock Gpt-4 technical report, 2023{\natexlab{c}}.

\bibitem[Peng et~al.(2023)Peng, Wang, Dong, Hao, Huang, Ma, and Wei]{peng2023kosmos}
Zhiliang Peng, Wenhui Wang, Li Dong, Yaru Hao, Shaohan Huang, Shuming Ma, and Furu Wei.
\newblock Kosmos-2: Grounding multimodal large language models to the world.
\newblock \emph{arXiv preprint arXiv:2306.14824}, 2023.

\bibitem[Plummer et~al.(2015)Plummer, Wang, Cervantes, Caicedo, Hockenmaier, and Lazebnik]{plummer2015flickr30k}
Bryan~A Plummer, Liwei Wang, Chris~M Cervantes, Juan~C Caicedo, Julia Hockenmaier, and Svetlana Lazebnik.
\newblock Flickr30k entities: Collecting region-to-phrase correspondences for richer image-to-sentence models.
\newblock In \emph{ICCV}, 2015.

\bibitem[Radford et~al.(2021)Radford, Kim, Hallacy, Ramesh, Goh, Agarwal, Sastry, Askell, Mishkin, Clark, et~al.]{CLIP}
Alec Radford, Jong~Wook Kim, Chris Hallacy, Aditya Ramesh, Gabriel Goh, Sandhini Agarwal, Girish Sastry, Amanda Askell, Pamela Mishkin, Jack Clark, et~al.
\newblock Learning transferable visual models from natural language supervision.
\newblock \emph{arXiv preprint arXiv:2103.00020}, 2021.

\bibitem[Touvron et~al.(2023)Touvron, Lavril, Izacard, Martinet, Lachaux, Lacroix, Rozi{\`e}re, Goyal, Hambro, Azhar, et~al.]{touvron2023llama}
Hugo Touvron, Thibaut Lavril, Gautier Izacard, Xavier Martinet, Marie-Anne Lachaux, Timoth{\'e}e Lacroix, Baptiste Rozi{\`e}re, Naman Goyal, Eric Hambro, Faisal Azhar, et~al.
\newblock Llama: Open and efficient foundation language models.
\newblock \emph{arXiv preprint arXiv:2302.13971}, 2023.

\bibitem[Wang et~al.(2023{\natexlab{a}})Wang, Chen, Chen, Wu, Zhu, Zeng, Luo, Lu, Zhou, Qiao, and Dai]{wang2023visionllm}
Wenhai Wang, Zhe Chen, Xiaokang Chen, Jiannan Wu, Xizhou Zhu, Gang Zeng, Ping Luo, Tong Lu, Jie Zhou, Yu Qiao, and Jifeng Dai.
\newblock Visionllm: Large language model is also an open-ended decoder for vision-centric tasks, 2023{\natexlab{a}}.

\bibitem[Wang et~al.(2023{\natexlab{b}})Wang, Lv, Yu, Hong, Qi, Wang, Ji, Yang, Zhao, Song, Xu, Xu, Li, Dong, Ding, and Tang]{wang2023cogvlm}
Weihan Wang, Qingsong Lv, Wenmeng Yu, Wenyi Hong, Ji Qi, Yan Wang, Junhui Ji, Zhuoyi Yang, Lei Zhao, Xixuan Song, Jiazheng Xu, Bin Xu, Juanzi Li, Yuxiao Dong, Ming Ding, and Jie Tang.
\newblock Cogvlm: Visual expert for pretrained language models.
\newblock 2023{\natexlab{b}}.

\bibitem[Yan et~al.(2023)Yan, Jiang, Wu, Wang, Yuan, Luo, and Lu]{UNINEXT}
Bin Yan, Yi Jiang, Jiannan Wu, Dong Wang, Zehuan Yuan, Ping Luo, and Huchuan Lu.
\newblock Universal instance perception as object discovery and retrieval.
\newblock In \emph{CVPR}, 2023.

\bibitem[Yang et~al.(2023)Yang, Zhang, Li, Zou, Li, and Gao]{yang2023setofmark}
Jianwei Yang, Hao Zhang, Feng Li, Xueyan Zou, Chunyuan Li, and Jianfeng Gao.
\newblock Set-of-mark prompting unleashes extraordinary visual grounding in gpt-4v, 2023.

\bibitem[Yang et~al.(2022)Yang, Gan, Wang, Hu, Ahmed, Liu, Lu, and Wang]{yang2022unitab}
Zhengyuan Yang, Zhe Gan, Jianfeng Wang, Xiaowei Hu, Faisal Ahmed, Zicheng Liu, Yumao Lu, and Lijuan Wang.
\newblock Unitab: Unifying text and box outputs for grounded vision-language modeling, 2022.

\bibitem[Yao et~al.(2022)Yao, Han, Wen, Liang, Xu, Zhang, Li, Xu, and Xu]{yao2022detclip}
Lewei Yao, Jianhua Han, Youpeng Wen, Xiaodan Liang, Dan Xu, Wei Zhang, Zhenguo Li, Chunjing Xu, and Hang Xu.
\newblock Detclip: Dictionary-enriched visual-concept paralleled pre-training for open-world detection, 2022.

\bibitem[Yao et~al.(2023)Yao, Han, Liang, Xu, Zhang, Li, and Xu]{yao2023detclipv2}
Lewei Yao, Jianhua Han, Xiaodan Liang, Dan Xu, Wei Zhang, Zhenguo Li, and Hang Xu.
\newblock Detclipv2: Scalable open-vocabulary object detection pre-training via word-region alignment, 2023.

\bibitem[Ye et~al.(2023)Ye, Hu, Xu, Ye, Yan, Dan, Zhao, Xu, Li, Tian, et~al.]{ye2023mplug_docowl}
Jiabo Ye, Anwen Hu, Haiyang Xu, Qinghao Ye, Ming Yan, Yuhao Dan, Chenlin Zhao, Guohai Xu, Chenliang Li, Junfeng Tian, et~al.
\newblock mplug-docowl: Modularized multimodal large language model for document understanding.
\newblock \emph{arXiv preprint arXiv:2307.02499}, 2023.

\bibitem[You et~al.(2023)You, Zhang, Gan, Du, Zhang, Wang, Cao, Chang, and Yang]{you2023ferret}
Haoxuan You, Haotian Zhang, Zhe Gan, Xianzhi Du, Bowen Zhang, Zirui Wang, Liangliang Cao, Shih-Fu Chang, and Yinfei Yang.
\newblock Ferret: Refer and ground anything anywhere at any granularity, 2023.

\bibitem[Yu et~al.(2016)Yu, Poirson, Yang, Berg, and Berg]{yu2016modeling}
Licheng Yu, Patrick Poirson, Shan Yang, Alexander~C. Berg, and Tamara~L. Berg.
\newblock Modeling context in referring expressions, 2016.

\bibitem[Zhang et~al.(2022{\natexlab{a}})Zhang, Li, Liu, Zhang, Su, Zhu, Ni, and Shum]{zhang2022dino}
Hao Zhang, Feng Li, Shilong Liu, Lei Zhang, Hang Su, Jun Zhu, Lionel~M. Ni, and Heung-Yeung Shum.
\newblock Dino: Detr with improved denoising anchor boxes for end-to-end object detection, 2022{\natexlab{a}}.

\bibitem[Zhang et~al.(2022{\natexlab{b}})Zhang, Zhang, Hu, Chen, Li, Dai, Wang, Yuan, Hwang, and Gao]{zhang2022glipv2}
Haotian* Zhang, Pengchuan* Zhang, Xiaowei Hu, Yen-Chun Chen, Liunian~Harold Li, Xiyang Dai, Lijuan Wang, Lu Yuan, Jenq-Neng Hwang, and Jianfeng Gao.
\newblock Glipv2: Unifying localization and vision-language understanding.
\newblock \emph{arXiv preprint arXiv:2206.05836}, 2022{\natexlab{b}}.

\bibitem[Zhang et~al.(2023{\natexlab{a}})Zhang, Li, Zou, Liu, Li, Gao, Yang, and Zhang]{zhang2023simple}
Hao Zhang, Feng Li, Xueyan Zou, Shilong Liu, Chunyuan Li, Jianfeng Gao, Jianwei Yang, and Lei Zhang.
\newblock A simple framework for open-vocabulary segmentation and detection.
\newblock \emph{arXiv preprint arXiv:2303.08131}, 2023{\natexlab{a}}.

\bibitem[Zhang et~al.(2023{\natexlab{b}})Zhang, Han, Liu, Gao, Zhou, Hu, Yan, Lu, Li, and Qiao]{zhang2023llamaadapter}
Renrui Zhang, Jiaming Han, Chris Liu, Peng Gao, Aojun Zhou, Xiangfei Hu, Shilin Yan, Pan Lu, Hongsheng Li, and Yu Qiao.
\newblock Llama-adapter: Efficient fine-tuning of language models with zero-init attention, 2023{\natexlab{b}}.

\bibitem[Zhang et~al.(2023{\natexlab{c}})Zhang, Sun, Chen, Xiao, Shao, Zhang, Chen, and Luo]{zhang2023gpt4roi}
Shilong Zhang, Peize Sun, Shoufa Chen, Min Xiao, Wenqi Shao, Wenwei Zhang, Kai Chen, and Ping Luo.
\newblock Gpt4roi: Instruction tuning large language model on region-of-interest.
\newblock \emph{arXiv preprint arXiv:2307.03601}, 2023{\natexlab{c}}.

\bibitem[Zhao et~al.(2023)Zhao, Lin, Zhou, Huang, Feng, and Kang]{zhao2023bubogpt}
Yang Zhao, Zhijie Lin, Daquan Zhou, Zilong Huang, Jiashi Feng, and Bingyi Kang.
\newblock Bubogpt: Enabling visual grounding in multi-modal llms.
\newblock \emph{arXiv preprint arXiv:2307.08581}, 2023.

\bibitem[Zhu et~al.(2022)Zhu, Zhou, Shen, Luo, Pan, Lin, Chen, Cao, Sun, and Ji]{zhu2022seqtr}
Chaoyang Zhu, Yiyi Zhou, Yunhang Shen, Gen Luo, Xingjia Pan, Mingbao Lin, Chao Chen, Liujuan Cao, Xiaoshuai Sun, and Rongrong Ji.
\newblock Seqtr: A simple yet universal network for visual grounding.
\newblock In \emph{Computer Vision--ECCV 2022: 17th European Conference, Tel Aviv, Israel, October 23--27, 2022, Proceedings, Part XXXV}, pages 598--615. Springer, 2022.

\bibitem[Zhu et~al.(2023)Zhu, Chen, Shen, Li, and Elhoseiny]{zhu2023minigpt}
Deyao Zhu, Jun Chen, Xiaoqian Shen, Xiang Li, and Mohamed Elhoseiny.
\newblock Minigpt-4: Enhancing vision-language understanding with advanced large language models.
\newblock \emph{arXiv preprint arXiv:2304.10592}, 2023.

\end{thebibliography}
}

\clearpage
\setcounter{page}{1}
\maketitlesupplementary

\appendix


\begin{figure}[t!]
\centering  
\vspace{-0mm}
\includegraphics[width=0.5\textwidth]{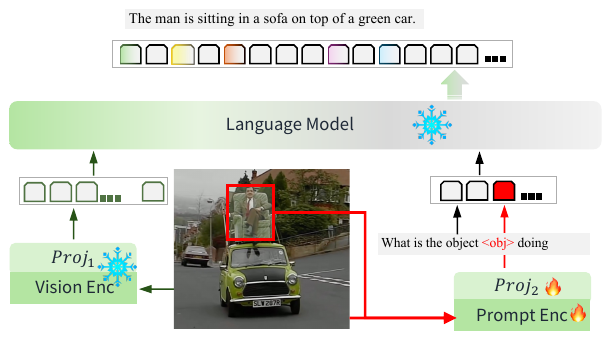} \\
\vspace{-2mm}
\caption{\footnotesize Network architecture of our \fullname{} for supporting visual prompts. Snow flake denotes the part is frozen and fire denotes the part is trainable.}
\label{fig: visual prompt arch}  
  \vspace{-3mm}
\end{figure}
\section{More details about visual prompt}
We support visual prompts such as clicks, boxes and scribbles with low training cost by only train the visual prompt encoder to align prompt features to the language space.\\

\noindent\textbf{Training.} We add the support of visual prompt seamlessly to the trained grounded visual chat model. We use a pretrained Semantic-SAM model as the prompt encoder. As shown in Fig.~\ref{fig: visual prompt arch}, a visual prompt will be encode as a prompt feature by prompt encoder and map to the language space with a projection layer. There is a special token $\langle obj \rangle$ in the text instruction, whose input embedding will be replaced by the visual prompt feature. To avoid any influence on the existing model, we propose to fix the other part of the model and only tune the prompt encoder and the projection layer of the prompt encoder.\\

\noindent\textbf{Data.} As demonstrated in Section~\ref{sec: data}, we created 150K grounded visual chat data with GPT4 to match none phrases in answers with GT instances. 
The data for visual prompts is annotated in the same way as the grounded visual chat data. As shown in the first blocks of Table~\ref{tab:full_example_car_bbox}, we simply change the sentence in Context 2 to questions and GPT4 can help to match entities in the question with GT instances. Then, we can sample various types of visual prompts based on the GT instances. An example data is as follows. Assume the input image is that in Fig.~\ref{fig: visual prompt arch} and the original question is \textit{Q: What is the man doing?}. With the help of GPT4, ``the man"  is matched to the red box in the image. Then we change the text prompt to \textit{Q: What is the object $\langle obj\rangle$ doing?}. $\langle obj\rangle$ is a place holder. Its input embedding will be replaced by the prompt embedding of ``the man". We labeled a part of 150K LLaVA instruction tuning data to visual prompt data. The data has high-quality visual chat because it is generated by GPT4. 

However, it is not good at distinguishing different instances in the image because it usually  talks about a few main objects for each image. For many images, the data only contain discussions about one instance. This is not good for the model to distinguish between different instances. Therefore, we include RefCOCO to train visual prompt encoder for distinguishing instances. RefCOCO has several instance, text pairs denoted as $(I,T)$ for each image. We convert them as visual chat data with the following template: \textit{Q:Please describe the object $\langle obj \rangle$ briefly. A: $T$} where $T$ is the referring text.

\section{Implementation details}
\begin{table}[]
\centering
\begin{tabular}{lll}
\toprule
                & stage1                    & stage2                    \\
                \midrule
$lr_{llm}$      & 0.0                       & $2e-5$                    \\
$lr_{gd}$       & $1e-4$                    & $1e-4$                    \\
$bs_{lang}$     & 32                        & 128                       \\
$bs_{gd}$       & 32                        & 64                        \\
warm up ratio   & 0.03                      & 0.03                      \\
weight decay    & 0.0                       & 0.0                       \\
bf16            & \checkmark & \checkmark \\
tf32            & \checkmark & \checkmark \\
grad accumulate & 2                         & 1                         \\
$n_{steps}$     & 10000                     & 8000                     \\
\bottomrule
\end{tabular}
\caption{The hyper-parmeters for stage1 and stage2.}
\label{tab: hyper-param}
\end{table}
We provide more details of our experiment configuration for reproducing our method. We provide hyper-parameters for both stage1 and stage2 in Table~\ref{tab: hyper-param}.
\begin{table*}[h!]\centering

\begin{minipage}{1.99\columnwidth}\vspace{0mm}    \centering
\begin{tcolorbox} 
    \centering
    \small
     \hspace{-6mm}
\begin{itemize}[leftmargin=7.5mm]
\setlength{\itemsep}{2pt}
\item "Please segment $\langle \mathrm{phrase} \rangle$."
\item "Can you segment $\langle \mathrm{phrase} \rangle$?"
\item "Please provide the boxes and masks for $\langle \mathrm{phrase} \rangle$."
\item "We need the boxes and masks for $\langle \mathrm{phrase} \rangle$, please."
\item "Ensure that the boxes and masks for $\langle \mathrm{phrase} \rangle$ are provided."
\item "Include the boxes and masks for $\langle \mathrm{phrase} \rangle$ in your submission."
\item "Don't forget to attach the boxes and masks for $\langle \mathrm{phrase} \rangle$."
\item "It's important to have the boxes and masks for $\langle \mathrm{phrase} \rangle$."
\item "Please remember to include $\langle \mathrm{phrase} \rangle$'s boxes and masks."
\item "The request is for the boxes and masks related to $\langle \mathrm{phrase} \rangle$."
\item "Kindly submit the boxes and masks corresponding to $\langle \mathrm{phrase} \rangle$."

\end{itemize}

\end{tcolorbox}
    
\vspace{-2mm}
\caption{The list of instruction templates for referring expression tasks. $\langle \mathrm{phrase} \rangle$ denotes the referring expression.}
    \label{tab:res}
\end{minipage}
\end{table*}

\begin{table*}[h!]\centering

\begin{minipage}{1.99\columnwidth}\vspace{0mm}    \centering
\begin{tcolorbox} 
    \centering
    \small
     \hspace{-6mm}
\begin{itemize}[leftmargin=7.5mm]
\setlength{\itemsep}{2pt}
\item "Describe the image concisely."
\item "Provide a brief description of the given image."
\item "Offer a succinct explanation of the picture presented."
\item "Summarize the visual content of the image."
\item "Give a short and clear explanation of the subsequent image."
\item "Share a concise interpretation of the image provided."
\item "Present a compact description of the photo's key features."
\item "Relay a brief, clear account of the picture shown."
\item "Render a clear and concise summary of the photo."
\item "Write a terse but informative summary of the picture."
\item "Create a compact narrative representing the image presented."
\end{itemize}

\end{tcolorbox}
    
\vspace{-2mm}
\caption{The list of instructions for brief image description.}
    \label{tab:short caption}
\end{minipage}
\end{table*}

\begin{table*}[h!]\centering

\begin{minipage}{1.99\columnwidth}\vspace{0mm}    \centering
\begin{tcolorbox} 
    \centering
    \small
     \hspace{-6mm}
\begin{itemize}[leftmargin=7.5mm]
\setlength{\itemsep}{2pt}
\item "with grounding"
\item "with boxes and masks"
\item "Please also provide the boxes and masks for the noun phrases in the response."
\item "Kindly ensure that the response includes the relevant boxes and masks for each noun phrase."
\item "Additionally, include the boxes and masks that match each noun phrase in the response."
\item "Please provide the boxes and masks that correspond to every noun phrase in your response."
\item "It's important to have the boxes and masks that align with each noun phrase in the response."
\item "Make sure to include the appropriate boxes and masks for each noun phrase in your response."
\item "In your response, include the boxes and masks that pertain to each noun phrase."
\item "Also, supply the boxes and masks that are linked to each noun phrase in the response."
\item "Additionally, please furnish the boxes and masks that correspond to each noun phrase in the response."
\item "Don't forget to provide the boxes and masks associated with each noun phrase in your response."
\item "Ensure that each noun phrase in the response has its respective boxes and masks."
\end{itemize}

\end{tcolorbox}
    
\vspace{-2mm}
\caption{The list of suffixes of the instructions for grounding caption.}
    \label{tab:grounded caption}
\end{minipage}
\end{table*}

\begin{table*}[h!]\centering

\begin{minipage}{1.99\columnwidth}\vspace{0mm}    \centering
\begin{tcolorbox} 
    \centering
    \small
     \hspace{-6mm}
\begin{itemize}[leftmargin=7.5mm]
\setlength{\itemsep}{2pt}
\item "Could you offer a concise description of the object $\langle obj \rangle$ ?"
\item "Please give a short summary of the object $\langle obj \rangle$ .",
\item "Kindly provide a succinct description of the object $\langle obj \rangle$ ."
\item "I'd appreciate a brief overview of the object $\langle obj \rangle$ ."
\item "Could you summarize the object $\langle obj \rangle$  in a few words?"
\item "Please provide a brief explanation of the object $\langle obj \rangle$ ."
\item "I'd like a quick description of the object $\langle obj \rangle$ , please."
\item "Could you quickly describe the object $\langle obj \rangle$ ?"
\item "Please offer a condensed description of the object $\langle obj \rangle$ ."
\item "Can you briefly characterize the object $\langle obj \rangle$ ?"
\item "Kindly provide a brief description of the object $\langle obj \rangle$ ."
\item "Would you mind giving a brief account of the object $\langle obj \rangle$ ?"
\item "Can you sketch a short description of the object $\langle obj \rangle$ ?"
\item "Please deliver a terse description of the object $\langle obj \rangle$ ."
\end{itemize}

\end{tcolorbox}
    
\vspace{-2mm}
\caption{The list of instructions for region caption data.}
    \label{tab:region caption}
\end{minipage}
\end{table*}

\section{Instruction lists for different data format.}
In this section, we give instruction lists used to construct instruction following data from different formats of data.

\section{More visualizations}
Table~\ref{tab:visual_example openset} and \ref{tab:visual_example coco} shows more comparison of \shortname{} and other grounding LMMs in grounded description. 
\begin{table*}
  \begin{minipage}{0.99\textwidth}
\centering  
\vspace{-4mm}
\scalebox{0.88}{
\begin{tabular}{l p{10.9cm} r }
\toprule
 \multicolumn{3}{l}{\bf Grounded description example 2:}  \\
\midrule
User & Describe the following image with grounding. &
\multirow{2}{*}{\includegraphics[height=4.0cm]{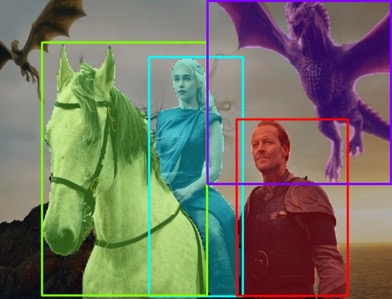}} \\
\shortname{} & The image features a fantasy scene with \textcolor{color_37951}{a woman} and \textcolor{color_274846}{a man} standing close together, riding on the back of \textcolor{color_159743}{horses}. The woman is on the left side of the image, and the man is on the right side. Both individuals appear to be enjoying their time on horseback, surrounded by a visually appealing fantasy setting. In the background, there are \textcolor{color_152093}{two dragons}, one on the left side and the other on the right side of the image. The dragons are situated near the top, adding a sense of depth and intrigue to the scene.
& \vspace{0.7cm}\\
\midrule
User & Can you provide a description of the image and include the coordinates [[x0,y0,x1,y1]] for each mentioned object?&
\multirow{2}{*}{ \includegraphics[height=4.0cm]{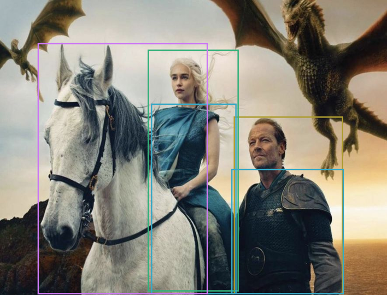}  }
\\
CogVLM-Grounding &  \textcolor{color_182125}{A man} and \textcolor{color_35381}{a woman}, both dressed in \textcolor{color_88082}{armor}, are riding \textcolor{color_229830}{horses}. & \vspace{2.9cm}\\
 \midrule
User & Provide a comprehensive description of the image <image> and specify the positions of any mentioned objects in square brackets.
&
\multirow{2}{*}{\includegraphics[height=4.0cm]{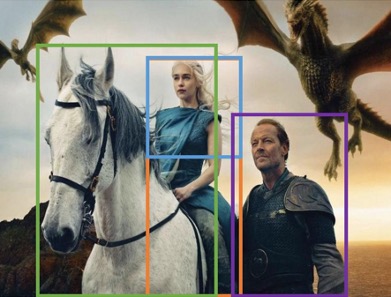} }
\\
Shikra & A \textcolor{color_261472}{woman} with \textcolor{color_122260}{blond-hair} sitting on a white \textcolor{color_142857}{horse} with a \textcolor{color_139071}{man} standing next to her. 
& \vspace{2.5cm} \\
\midrule
User & [grounding] Describe the scene in detail. &
\multirow{2}{*}{ \includegraphics[height=4.0cm]{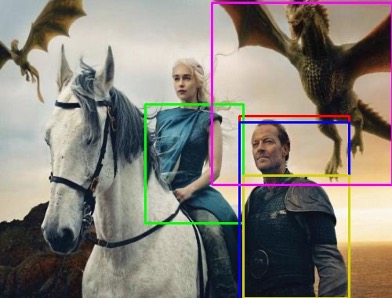} }
\\
MiniGPT v2 & \textcolor{color_274846}{A woman} in \textcolor{color_37696}{blue} is riding a white horse ,\textcolor{color_30046}{a man} in \textcolor{color_238111}{black} is walking towards the horse , and \textcolor{color_275101}{a dragon} is in the background
&  \vspace{2.9cm} \\
\bottomrule
\end{tabular}
}
\captionof{table}{An example of grounded detailed descriptions generated by \shortname{} and other LMMs. We use the best prompt recommended by the paper for each model.}  
\label{tab:visual_example openset}  
  \end{minipage}
\end{table*}
\begin{table*}
  \begin{minipage}{0.99\textwidth}
\centering  
\vspace{-4mm}
\scalebox{0.88}{
\begin{tabular}{l p{10.9cm} r }
\toprule
 \multicolumn{3}{l}{\bf Grounded description example 3:}  \\
\midrule
User & Describe the following image with grounding. & 
\multirow{2}{*}{\includegraphics[height=4.0cm]{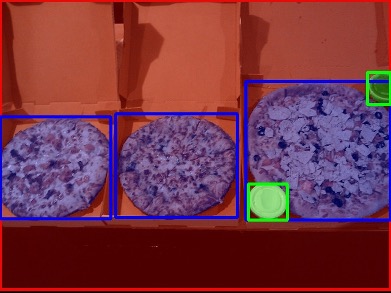}}
\\
\shortname{} & The image features \textcolor{cell_phone}{a dining table} with \textcolor{bottle}{three pizzas} on separate open boxes placed in the center. The pizzas have various toppings, making them visually appealing. In addition to the pizzas, there are \textcolor{TV}{two bowls} on the table, one located near the top left corner and the other at the top right. The table appears to be set for a meal, with the pizzas and bowls ready for guests to enjoy.
& \vspace{1.6cm} \\
\midrule
User & Can you provide a description of the image and include the coordinates [[x0,y0,x1,y1]] for each mentioned object?&
\multirow{2}{*}{ \includegraphics[height=4.0cm]{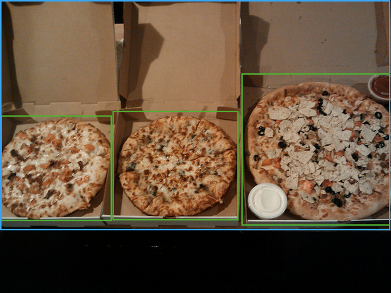}  }
\\
CogVLM-Grounding &  Three \textcolor{color_182125}{pizzas} in a \textcolor{color_91812}{box}.
& \vspace{2.85cm}\\
 \midrule
User & Provide a comprehensive description of the image <image> and specify the positions of any mentioned objects in square brackets.
&
\multirow{2}{*}{\includegraphics[height=4.0cm]{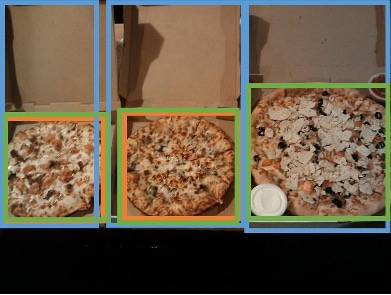} }
 \\
Shikra & Three \textcolor{color_261472}{pizzas} in pizza \textcolor{color_122260}{boxes} with various \textcolor{color_142857}{toppings}.
& \vspace{2.8cm} \\
\midrule
User & [grounding] Describe the scene in detail.  & 
\multirow{2}{*}{ \includegraphics[height=4.0cm]{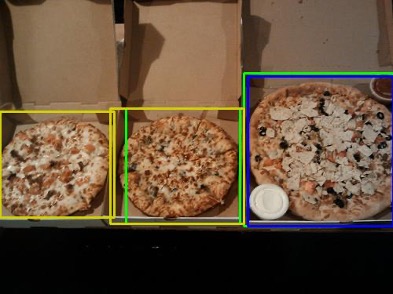} }
\\
MiniGPT v2 & \textcolor{color_274846}{Three pizza boxes} with \textcolor{color_37696}{pizzas} and \textcolor{color_30046}{toppings} on \textcolor{color_238111}{them}
&  \vspace{3.25cm} \\
\bottomrule
\end{tabular}
}
\captionof{table}{An example of grounded detailed descriptions generated by \shortname{} and other LMMs. We use the best prompt recommended by the paper for each model.}  
\label{tab:visual_example coco}  
  \end{minipage}
\end{table*}

\end{document}